\documentclass{article}

\usepackage[final, nonatbib]{neurips_2025}
\usepackage[numbers,sort&compress]{natbib}
 
\usepackage[utf8]{inputenc} % allow utf-8 input
\usepackage[T1]{fontenc}    % use 8-bit T1 fonts
\usepackage{hyperref}       % hyperlinks
\usepackage{url}            % simple URL typesetting
\usepackage{booktabs}       % professional-quality tables
\usepackage{amsfonts}       % blackboard math symbols
\usepackage{nicefrac}       % compact symbols for 1/2, etc.
\usepackage{microtype}      % microtypography
\usepackage[pdftex]{graphicx}
\usepackage{amsmath}
\usepackage{verbatim}
\usepackage{enumitem}
\usepackage[table]{xcolor}
\usepackage{wrapfig}
\usepackage{float}
\usepackage{nameref}
\usepackage{adjustbox}  % 放在导言区
\usepackage{multirow}

\usepackage{colortbl}
\usepackage{etoolbox}
\makeatletter
\patchcmd{\CT@row@color}{\global\let\CT@row@color@old\CT@row@color}{\CT@row@color@old}{}{}
\makeatother

\usepackage[framemethod=TikZ]{mdframed}  
\newcounter{theo}
\renewcommand{\thetheo}{}
\newenvironment{theo}[2][]{%
\refstepcounter{theo}%
\ifstrempty{#1}%
{\mdfsetup{%
frametitle={%
\tikz[baseline=(current bounding box.east),outer sep=0pt]
\node[anchor=east,rectangle,fill=blue!20]
{\strut \thetheo~#1};}}%
}%
\mdfsetup{innertopmargin=5pt,linecolor=cyan!25,% or blue!20
linewidth=2pt,topline=true,% 
roundcorner=5pt,
frametitleaboveskip=\dimexpr-\ht\strutbox\relax
}
\begin{mdframed}[]\relax%
\label{#2}}{\end{mdframed}}
\newcommand{\err}[2]{\ensuremath{#1\,{\scriptscriptstyle\pm #2}}}
\newcommand{\berr}[2]{\ensuremath{\mathbf{#1}\,{\scriptscriptstyle\pm \mathbf{#2}}}}

\hypersetup{
    colorlinks=true,
    linkcolor=red,
    citecolor=cyan,
    filecolor=magenta,      
    urlcolor=magenta,
}

\title{Neural-Driven Image Editing}

\vspace{-5mm}

\author{
\hspace{-0.1cm}
    \textbf{Pengfei Zhou\textsuperscript{1*$\dagger$},
    Jie Xia\textsuperscript{2,8*$\dagger$},
    Xiaopeng Peng\textsuperscript{3$\dagger$},
    Wangbo Zhao\textsuperscript{1},
    Zilong Ye\textsuperscript{2,8},
    Zekai Li\textsuperscript{1},} \\
\hspace{-0.2cm}
    \textbf{Suorong Yang\textsuperscript{1,4},
    Jiadong Pan\textsuperscript{2,8},
    Yuanxiang Chen\textsuperscript{5},
    Ziqiao Wang\textsuperscript{1},
    Kai Wang\textsuperscript{1},
    Qian Zheng\textsuperscript{2},} \\
\hspace{-0.1cm}
    \textbf{Xiaojun Chang\textsuperscript{5},
    Hao Jin\textsuperscript{2},
    Gang Pan\textsuperscript{2},
    Shurong Dong\textsuperscript{2$\ddagger$},
    Kaipeng Zhang\textsuperscript{6,7$\ddagger$},
    Yang You\textsuperscript{1}} \\
\textsuperscript{1}NUS 
\textsuperscript{2}ZJU
\textsuperscript{3}RIT
\textsuperscript{4}NJU
\textsuperscript{5}USTC
\textsuperscript{6}Shanghai AI Lab
\textsuperscript{7}SII
\textsuperscript{8}Hangzhou RongNao Tech \\
zpf4wp@outlook.com, \{jiexia, dongshurong\}@zju.edu.cn, zhangkaipeng@pjlab.org.cn\\
\thanks{Equal contribution; 
 $^\dagger$Core contributor;
 $^\ddagger$Corresponding author.}

}

\begin{document}

\maketitle

\vspace{-10mm}

\begin{figure*}[ht]
  \centering
  \includegraphics[width=0.95\textwidth]{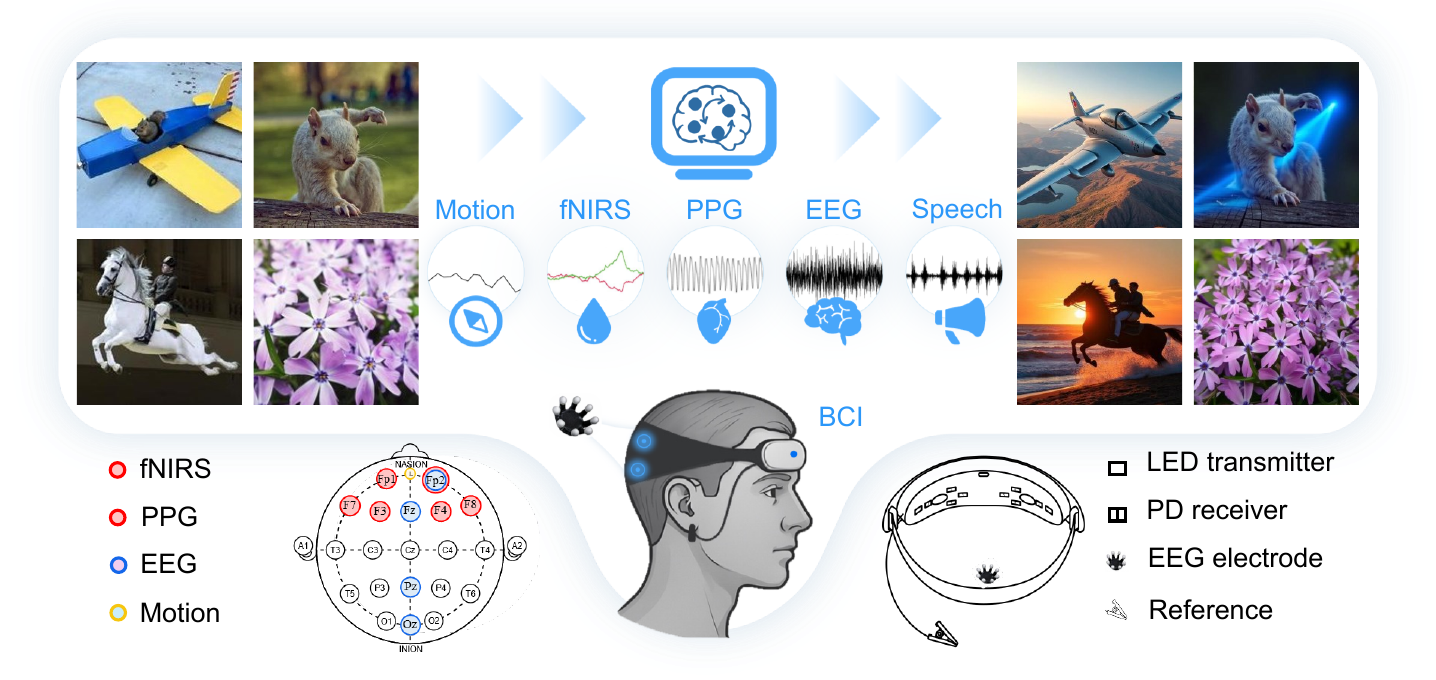}
  \vspace{-0.5cm}
  \caption{Illustration of LoongX for hands-free image editing via multimodal neural signals.}
  \label{fig:teaser}
\end{figure*}

\begin{abstract}

Traditional image editing typically relies on manual prompting, making it labor-intensive and inaccessible to individuals with limited motor control or language abilities. Leveraging recent advances in brain-computer interfaces (BCIs) and generative models, we propose \textbf{LoongX}, a hands-free image editing approach driven by multimodal neurophysiological signals. 
LoongX utilizes state-of-the-art diffusion models trained on a comprehensive dataset of 23,928 image editing pairs, each paired with synchronized electroencephalography (EEG), functional near-infrared spectroscopy (fNIRS), photoplethysmography (PPG), and head motion signals that capture user intent.
To effectively address the heterogeneity of these signals, LoongX integrates two key modules. The cross-scale state space (CS3) module encodes informative modality-specific features. The dynamic gated fusion (DGF) module further aggregates these features into a unified latent space, which is then aligned with edit semantics via fine-tuning on a diffusion transformer (DiT).
Additionally, we pre-train the encoders using contrastive learning to align cognitive states with semantic intentions from embedded natural language.
Extensive experiments demonstrate that LoongX achieves performance comparable to text-driven methods (CLIP-I: 0.6605 vs. 0.6558; DINO: 0.4812 vs. 0.4636) and outperforms them when neural signals are combined with speech (CLIP-T: 0.2588 vs. 0.2549). These results highlight the promise of neural-driven generative models in enabling accessible, intuitive image editing and open new directions for cognitive-driven creative technologies. The code and dataset are released on the project website: \url{https://loongx1.github.io}.

\end{abstract}

\section{Introduction}
\label{intro}

Image editing involves manipulating digital visuals to achieve desired effects, significantly impacting fields like advertising, entertainment, and scientific visualization~\citep{brooks2023instructpix2pix}. Traditionally, this task demands extensive manual effort and technical expertise. Advances in generative models have streamlined instruction-based image editing through automated pipelines~\citep{zhao2024ultraedit, wang2024diffusion, openai2025gpt4o}. Nevertheless, these methods still heavily depend on intensive user inputs, such as text prompts~\citep{hertzprompt, sheynin2024emu}, visual references like masks or sketches~\citep{chang2023muse,zeng2022sketchedit}, and physical operations like dragging~\citep{shi2024dragdiffusion, shin2024instantdrag, liu2024magicquill, jiangclipdrag}. Such reliance limits efficiency and accessibility, especially for users with motor or communication impairments.

% Recent advances in brain-computer interfaces (BCIs) are opening up new possibilities for human-AI interaction~\citep{han2024onellm, allen2022massive, de2024cross}. 
To address these challenges, alternative input modalities have been explored~\citep{shuai2024survey, jiang2021talk, yang2024align} for image editing. Among these, brain–computer interfaces (BCIs) provide a promising possibility with their recent advancement in hardware precision~\citep{han2024onellm, allen2022massive}. Starting from early attempts in passive tasks such as mental state recognition~\citep{wang2024research} and neural activity analysis~\citep{duan2023dewave, benchetritbrain}, BCIs have begun to be involved in more active generative tasks such as neural-driven chat~\citep{baradari2025neurochat} and visual content creation~\citep{bai2024dreamdiffusion, liu2024eeg2video}.

However, existing approaches remain limited to the use of single-modality data such as electroencephalography (EEG)~\citep{Davis_2022_CVPR,bai2024dreamdiffusion} or functional magnetic resonance imaging (fMRI)~\citep{huo2024neuropictor}, which is insufficient to capture nuanced user intentions for enabling complex editing tasks. In practice, physiological signals from different modalities can offer complementary insights into cognitive states such as attention, motivation, and emotional regulation~\citep{Cai2025, Brake2024, Li2024,6815207, Si2023, Arsalan2021,Tazarv2021}, underscoring the need for multimodal neural information integration. 

Given the limited exploration of this emerging area, here we ask three key research questions (RQs):

\begin{enumerate}[label=\textbf{RQ\arabic*.},itemsep=0.5pt, parsep=0.5pt, topsep=-2pt]
\label{key:rqs}
    \item \label{key:rq1} Can neural signals alone drive instruction-based image editing?
    \item \label{key:rq2} If yes, what kind of information do multimodal neural signals contribute?
    \item \label{key:rq3} How do neural-signal conditions compare and complement natural-language instructions?
\end{enumerate}

To answer these questions, we construct \textbf{L-Mind}, a comprehensive multimodal dataset comprising 23,928 image pairs with synchronously collected EEG, functional near infrared spectroscopy (fNIRS)~\citep{cao2021brain}, photoplethysmography (PPG)~\citep{han2020development}, head motion, and speech signals from 12 participants conceiving image editing tasks. Captured using a wireless, lightweight BCI system that supports unconstrained head movements and speech~\citep{EEGfNIRS}, L-Mind offers higher ecological validity under natural real-world conditions and supports robust training of brain-supervised generative models.

% However, these modalities may not naturally provide structured features aligned with instruction intention. To enable effective feature extraction from vast signals with different shapes and facilitate deep multimodal integration, we propose two novel modules in LoongX: the cross-scale state space (CS3) encoder and dynamic gated fusion (DGF). The CS3 encoder is proposed to extract relevant neural and physiological features, and map them into precise editing instruction semantics, To support deep multimodal integration, we introduce the DAUF, which effectively combines neural signals from diverse sources and explores inter-modality interactions for more accurate semantic comprehension. We also pretrain the encoders with data combined from large-scale datasets ~\citep{nieto2022thinking} and our proposed L-Mind for aligning the extracted features with semantic information in text embeddings.

Building on L-Mind, we propose \textbf{LoongX}, a hands-free image editing approach that innovatively integrates the proposed multimodal neural signal fusion strategy with a diffusion transformer (DiT) to translate neural intent into image edits. Unlike prior single-modal methods, LoongX integrates EEG, fNIRS, PPG, and head motion signals, extracting explicit user intentions from EEG signals across multiple scalp regions, incorporating cognitive load and emotional valence data from fNIRS, and capturing stress and engagement indicators through PPG and motion signals. We introduce two new modules to manage diverse multimodal input: a cross-scale state space (CS3) encoder for robust feature extraction and a dynamic gated fusion (DGF) module for comprehensive multimodal integration. These encoders are pretrained via contrastive learning on combined large-scale datasets and L-Mind to align neural features with semantic text embeddings.

Extensive experiments qualitatively and quantitatively demonstrate the feasibility of neural-driven image editing. Integrated multimodal neural signals achieve performance comparable to text-driven baselines (CLIP-I: 0.6605 vs. 0.6558; DINO: 0.4812 vs. 0.4637). Combined neural signals with speech instructions surpass text prompts alone (CLIP-T: 0.2588 vs. 0.2549). Ablation studies verify the effectiveness of proposed modules and further explore the contribution of each signal, showing that EEG + fNIRS contribute most among signals, and the Oz and Fpz sites, as EEG input channels, represent the key brain region. These findings underscore LoongX’s potential to facilitate intuitive, inclusive image editing and inspire future human–AI interaction.

% EEG + fNIRS closes 90\% of the gap to text prompts, Pz-focused attention lifts CLIP-I by 1\%, and cross-modal pre-training halves the data required to reach text-level quality. 

Our main contributions are summarized as follows:
\begin{enumerate}[label=\arabic*), leftmargin=2.3em,itemsep=0.5pt, parsep=0.5pt, topsep=-2pt]
    \item \textbf{L-Mind}, a multimodal dataset with 23,928 image-editing pairs featuring synchronized EEG, fNIRS, PPG, motion, and speech signals collected in natural settings.
    \item \textbf{LoongX}, a novel neural-driven editing method with CS3 and DGF modules for effective feature extraction and multimodal integration (see the effect in Fig.~\ref{fig:teaser}).
    \item Extensive experiments validate multimodal neural signals' effectiveness and provide insights into modality-specific contributions and their synergy with speech-based inputs.
\end{enumerate}

\section{Related Works}
\label{related}

% \begin{figure}[t]
% 	\centering
%   \includegraphics[trim= 0.6cm 0.5cm 0 0, width=0.8\textwidth]{mechanism.pdf}
%   \caption{The principle of neural signal sensing. EEG data are obtained by measuring electrical signals from neurons; fNIRS and PPG data are obtained by measuring hemodynamics using infrared light.}
%   \vspace{-0.5cm}
%   \label{fig:collect} 
% \end{figure}

\subsection{Brain-supervised Generation}

As an emerging technology, brain–computer interface (BCI) builds direct communication between the brain and devices by decoding neural signals~\citep{Edelman2025Noninvasive, Eldawlatly2024GenerativeBCI}. Advances in machine learning have improved its accuracy, enabling brain-guided generative methods for visual content creation. Several recent methods integrate neurophysiological data (e.g., fMRI, EEG, or fNIRS) with generative models~\citep{Takagi_2023_CVPR, bai2024dreamdiffusion, tripathy2021decoding}. For instance, CMVDM aligns fMRI features with semantics for image synthesis~\citep{zeng2024controllable}, and the MindEye series further lifts the resolution of generated images from decoded fMRI~\citep{Scotti_2023_NeurIPS,Scotti_2024_ICML}. OneLLM leverages the large fMIR dataset for multimodal alignment in a multimodal large language model~\citep{han2024onellm}. DreamDiffusion produces images from EEG via temporal masked modeling~\citep{bai2024dreamdiffusion}. EEG2Video extends the idea to dynamic video content~\citep{Liu_2024_NeurIPS}. While Davis~\textit{et al.}~\citep{Davis_2022_CVPR} initially explore brain-guided semantic image editing using a generative adversarial network, this work is limited to facial images and EEG signals. Moreover, Adamic~\textit{et al.}~\citep{adamic2024progress} reconstructs visual images from brain activity measured by fNIRS.

Unlike previous studies, our data are collected using a wireless BCI system (Fig.~\ref{fig:collect}) as participants conceive instruction-based image edits. Compared with fMRI methods, our framework combines lightweight EEG, fNIRS, PPG, and head-motion signals, which can support greater portability and broader real-world applicability. To the best of our knowledge, this is the first work to fully leverage all these signals for instruction-based image editing, focusing on improved neural feature encoding and optimized multimodal fusion strategies.

\subsection{Instruction-based Image Editing}

Recent generative models like GPT-4o~\citep{gpt4} and Gemini~\citep{team2023gemini} have evolved from basic question answering to advanced image editing by interpreting user instructions. Modern instruction-based image editing agents integrate multimodal inputs, including text, images, and videos, to accurately identify and apply visual edits~\cite{shuai2024survey}. Leveraging learned multimodal representations, these agents interpret instructions from input, localize relevant regions, and perform targeted modifications~\cite{zhang2025context, tan2024ominicontrol, tan2025ominicontrol2}. Recent approaches, such as InstructPix2Pix~\citep{brooks2023instructpix2pix}, UltraEdit~\citep{zhao2024ultraedit}, MagicBrush~\citep{zhang2023magicbrush}, MIGE~\citep{tian2025mige}, and ACE~\citep{mao2025ace++} improve region-specific edits guided by natural language prompts. Speech-driven image editing \citep{jiang2021talk} was also explored, highlighting the feasibility of hands-free interaction but still limited by linguistic expressiveness in recorded speech.

Despite these advancements, achieving efficient, delicate, prompt-free image editing remains challenging. Our work addresses this gap, exploring neural-signal-driven editing agents to decode cognitive intent directly for image manipulation. It is believed that the findings in this work can significantly enhance accessibility and interaction efficiency in future BCI-enabled applications, particularly benefiting individuals with physical disabilities.

% 后面会加更多引用

% \vspace{-0.5em}

\section{Dataset}
\label{dataset}

\subsection{Data Collection}

\begin{figure*}[t]
	\centering
    \vspace{-1em}
  \includegraphics[trim= 0cm 0cm 0cm 0cm, clip, width=0.98\textwidth]{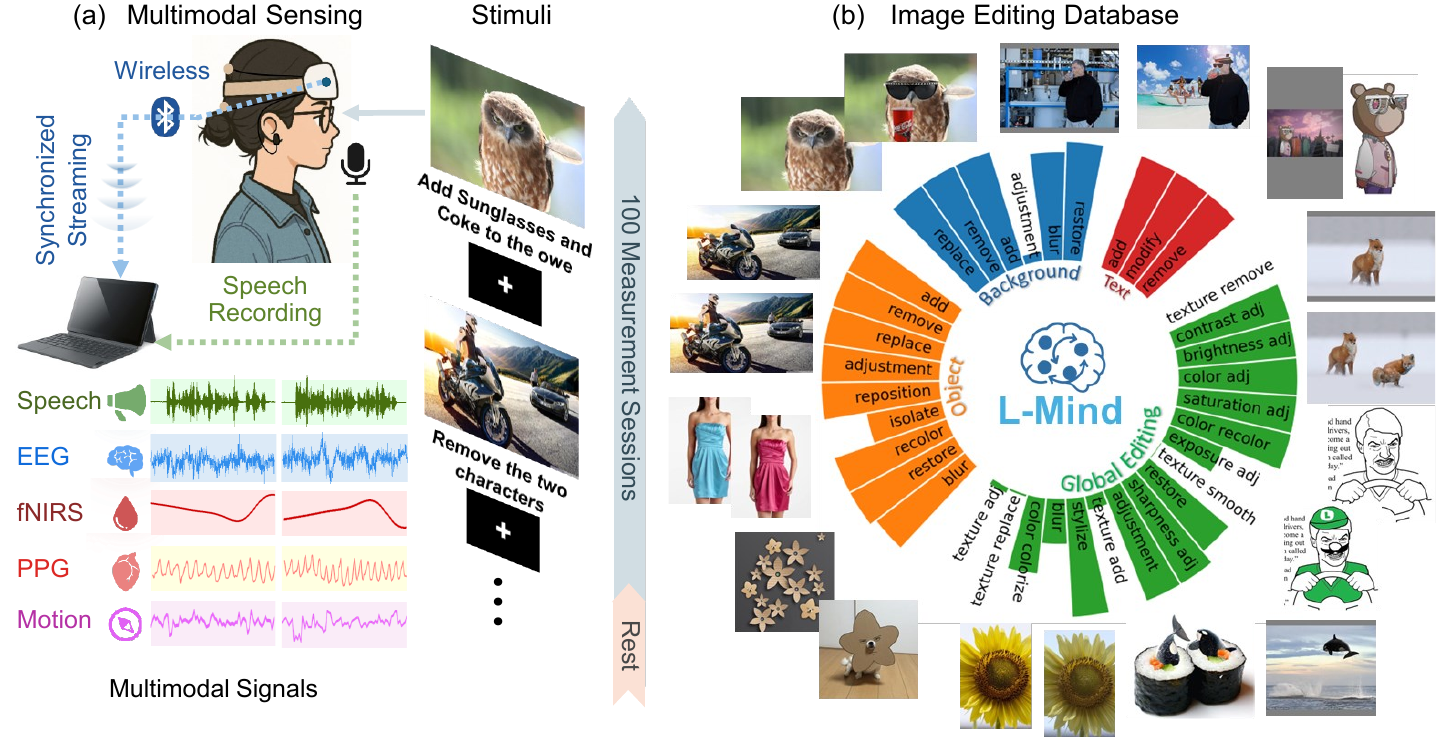}
  %\vspace{-0.2cm}
  \caption{The L-Mind dataset comprises 23,928 multimodal editing samples, each including an original image, a ground truth text editing instruction, a ground truth edited image, as well as measured EEG, fNIRS, PPG, motion and speech signals. (a) Multimodal data collection pipeline; (b) Illustration and statistics of 35 types of image editing tasks.}
  \vspace{-0.5cm}
  \label{fig:collect} 
\end{figure*}

We collect 23,928 editing samples (22,728 training, 1,200 testing) from 12 participants using the setup depicted in Fig.~\ref{fig:collect}. Participants wear our multimodal sensor while viewing image-text pairs sourced from SEED-Data-Edit \cite{ge2024seed} on a 25-inch monitor (resolution: $1980 \times 1080$). The measured EEG, fNIRS, and PPG physiological signals are streamed in real-time via Bluetooth 5.3, synchronized and aligned via lab streaming layer by the proprietary \textit{Lab Recorder} software~\citep{Kothe2023}. Participants simultaneously read displayed editing instructions aloud, providing audio speech signals.

Experiments are conducted in a quiet, temperature-controlled room (24°C, consistent humidity), starting at 9 AM daily. EEG signals are collected via non-invasive hydrogel electrodes, replaced every five hours to maintain signal quality. The experimental room is shielded from sunlight to prevent interference with fNIRS and PPG signals. Sessions start and end with participant-controlled audio recording and are marked by image names. Data from inactive intervals are excluded.

Each session (Fig.~\ref{fig:collect}) starts and ends with user-initiated audio recording and is labeled by the paired image. A 1-second cross-fixation follows each image pair, with breaks every 100 images. Twelve healthy college students (6 female, 6 male; mean age: 24.5 ± 2.5 years old) with normal or corrected vision participated. All participants gave informed consent and received financial compensation. The study was officially approved by the corresponding institute’s ethics committee.

\subsection{Data Preprocessing}

\textbf{EEG.} Four EEG channels (Pz, Fp2, Fpz, Oz; sampled at 250 Hz) undergo band-pass filtering (1–80 Hz) and notch filtering (48–52 Hz) to remove drifts, noise, and powerline interference. Ocular artifacts in Fp2 and Fpz are retained to capture eye movements.

\textbf{fNIRS.} Six-channel fNIRS signals (735 nm, 850 nm) are converted to relative hemoglobin concentration changes (HbO, HbR, HbT) using the Modified Beer–Lambert law. Optical density change is computed as $\Delta A(\lambda) = \log\left(I_0(\lambda)/I(\lambda) \right)$. Concentration changes are calculated as:
%\vspace{-1em}
\begin{equation}
\begin{bmatrix}
\Delta \text{HbO} \\
\Delta \text{HbR}
\end{bmatrix}
= \frac{1}{\text{DPF} \cdot L}
\cdot
\begin{bmatrix}
\varepsilon_{\text{HbO}}^{\lambda_1} & \varepsilon_{\text{HbR}}^{\lambda_1} \\
\varepsilon_{\text{HbO}}^{\lambda_2} & \varepsilon_{\text{HbR}}^{\lambda_2}
\end{bmatrix}^{-1}
\cdot
\begin{bmatrix}
\Delta A(\lambda_1) \\
\Delta A(\lambda_2)
\end{bmatrix}
\end{equation}
Hemodynamic signals ($\Delta$HbO, $\Delta$HbR, and $\Delta$HbT, where $\Delta$HbT = $\Delta$HbO + $\Delta$HbR) are band-pass filtered (0.01–0.5 Hz) to isolate relevant neural responses, averaged per hemisphere to reflect task-related brain activity.

\textbf{PPG and motion} Four-channel PPG signals (735 nm, 850 nm) are averaged per hemisphere via adaptive average pooling and filtered (0.5–4 Hz) to extract cardiac-related hemodynamic signals that reflect heart rate variability. Motion data from a six-axis sensor (12.5 Hz), capturing triaxial linear acceleration and angular velocity, characterizes head movements. See supplement for more details.
% More details can be found in the supplementary materials.

% This structured pipeline results in the high-quality L-Mind dataset, suitable for robust neural-driven image editing training and evaluation. More details can be found in the supplementary materials.

\vspace{-0.5em}

\section{Method}
\label{method}

\begin{figure*}[t]
	\centering
    \vspace{-0.5em}
	\includegraphics[trim=0cm 0cm 0 0, width=0.98\textwidth]{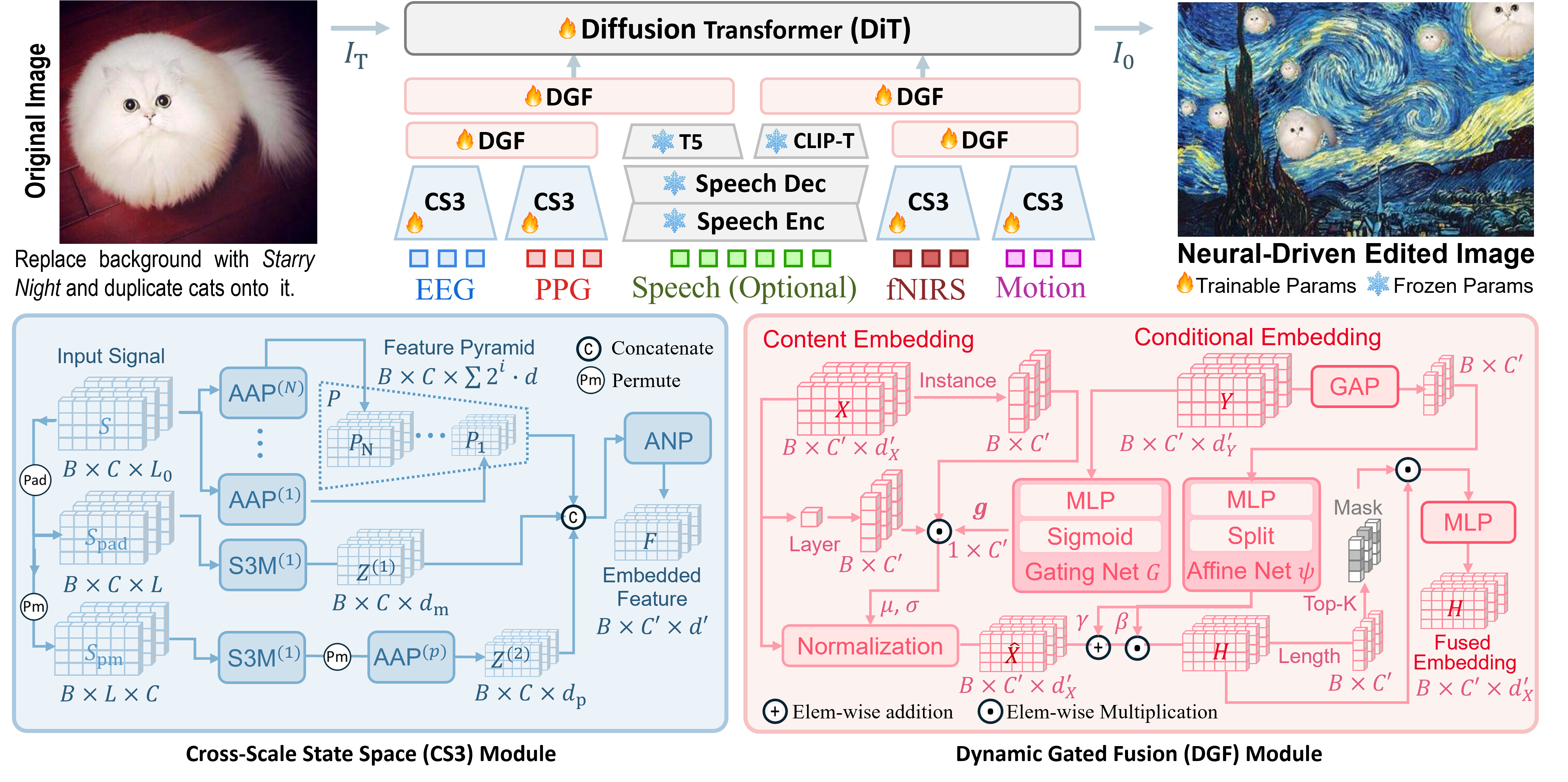}
    \vspace{-0.3cm}
 \caption{Overview of our proposed LoongX method for hands-free image editing. Receiving an input image, LoongX outputs an edited image using neural signals (and optional speech) as conditions.}
	\label{fig:model}
\vspace{-0.5em}
\end{figure*}

As illustrated in Fig.~\ref{fig:model}, LoongX extracts multimodal features from diverse neural signals and fuses them into a shared latent space in a pair-wise manner. Using Diffusion Transformer (DiT), the original image is translated into an edited image conditioned on the fused features. Following three research questions, we conduct a multi-label classification experiment (Sec.~\ref{subsec:classification_experiment}) showing that EEG outperforms noise by 20\%, and fusing all signals yields the highest F1 score. Combining neural signals with text achieves the best mAP, confirming modality complementarity. An input length of 8,192 gives the best performance with higher computational cost, motivating our framework’s design: a cross-scale state-space encoder for long sequences and dynamic gated fusion for feature integration.

\subsection{Cross-Scale State Space Encoding}
\label{subsec:cs3}

CS3 encoder extracts multi-scale features from diverse signals using an adaptive feature pyramid. To further capture dynamic spatio-temporal patterns beyond the fixed pyramid, CS3 uses a structured state space model (S3M) ~\cite{guefficiently} for efficient long-sequence encoding with linear complexity. To manage cost, it uses a cross-feature mechanism that separately encodes temporal and channel information.

\paragraph{Pyramid Encoding.} A single modality input signal $\mathbf S \in \mathbb{R}^{C\times L_0}$ is fed into an $N$-layer adaptive average pooling (AAP) module:
\begin{align}
  \{\mathbf{P}_i | i = 1,...,N\}=\mathrm{AAP}^{(i)}_{L\!\rightarrow\!s_i}(\mathbf S),
  \quad
  s_i=d\cdot2^{i} 
\end{align}
\noindent where we set $N=5$ and $d=64$ for EEG. The extracted embedding is computed as the concatenation of the feature pyramid $\mathbf P=\operatorname{Concat}(\{\mathbf{P}_i\})$.

\paragraph{State Space Encoding.}
To fully exploit both temporal and channel-wise dependencies in neural signals, we design a cross-shaped spatiotemporal encoding scheme, where one axis focuses on temporal patterns and the other on channel-wise dynamics. 

Specifically, the input signal $\mathbf{S}$ is padded from length $L_0$ to $L$, where $\mathbf{S}_\mathrm{pad}\in \mathbb{R}^{C\times L}$ with signal intensity normalized to $[-1,1]$. The padded signals and its permuted version $\mathbf{S}_\mathrm{pm}\in \mathbb{R}^{L\times C}$ are passed to two parallel S3M blocks, $\mathrm{S3M}^{(1)}$ and $\mathrm{S3M}^{(2)}$, respectively:
\begin{equation}
\mathbf Z^{(1)} = \mathrm{S3M}^{(1)}(\mathbf{S}_\mathrm{pad}),\quad \mathbf Z^{(2)} = \mathrm{S3M}^{(2)}(\mathbf{S}_\mathrm{pm}),
\label{eq:s3m-parallel}
\end{equation}
where each S3M block uses the continuous-time diagonal state-space model:
\begin{equation}
\dot{\mathbf e}(t)=\hat{\mathbf A}\mathbf e(t)+\hat{\mathbf B}s(t),\quad \mathbf z(t)=\hat{\mathbf C}\mathbf e(t)+\hat{\mathbf D}s(t),
\end{equation}
where $\mathbf{e}(t)$ denotes the latent state at time $t$, and $\hat{\mathbf{A}}$, $\hat{\mathbf{B}}$, $\hat{\mathbf{C}}$, $\hat{\mathbf{D}}$ are diagonal matrices that parameterize state transitions, input injection, state-to-output mapping, and direct input-to-output mapping, respectively. Due to the diagonal parameterization, the S3M block admits efficient computation with linear complexity $\mathcal O(L\log L)$. Through the S3M blocks, $\mathbf Z^{(1)}$ is down-sampled from length $L$ to $d_m$, yielding $\tilde{\mathbf Z}^{(1)}\!\in\!\mathbb{R}^{C\times d_{m}}$; $\mathbf Z^{(2)}$ is permuted and down-sampled via an AAP, giving $\tilde{\mathbf Z}^{(2)}\!\in\!\mathbb{R}^{C\times d_p}$.

\paragraph{Cross-Pyramid Aggregation.}

The encoder merges multi-scale and temporal streams along the channel dimension, resulting in:
\begin{align}
\mathbf F = \mathrm{ANP}\big(\operatorname{cat}_c(\tilde{\mathbf Z}^{(1)},\mathbf P,\tilde{\mathbf Z}^{(2)})\in\mathbb R^{C\times d'}\big).
\end{align}
where $d'=d_m + d_p + \Sigma^{N}_{i}2^i\cdot d$. The concatenated feature was projected via Adaptive Nonlinear Projection (ANP), which consists of two fully-connected layers, LayerNorm, ReLU, and dropout. The final embeded feature $\mathbf F\in\mathbb R^{C'\times L'}$ is obtained. 
% See supplementary materials for the values $d'$, $C$, $L$ that are associated with different signal modalities.

\subsection{Dynamic Gated Multimodal Fusion}
\label{subsec:DGF}
We propose the Dynamic Gated Fusion (DGF) module to dynamically bind a pair of content and condition embeddings to a unified latent space, which is further aligned with text embeddings. Our DGF includes gate mixing, adaptive affine modulation, and a dynamic masking block.

\paragraph{Gated Mixing.} We calculate instance-wise and layer-wise mean $\mu$ and variance $\sigma$ from input \textbf{content} embedding (e.g., EEG)
$\mathbf X\!\in\!\mathbb R^{C'\times L'_{X}}$ and \textbf{condition} embedding (e.g., PPG)
$\mathbf Y\!\in\!\mathbb R^{C'\times L'_{Y}}$ for further fusion into $\tilde{\mathbf H}\!\in\!\mathbb R^{C'\times L'_{X}}$ while emphasising informative channels and suppressing noise:

\begin{equation}
\underbrace{\begin{bmatrix}
\mu_{\text{inst}} & \sigma_{\text{inst}} \\[4pt]
\mu_{\text{layer}} & \sigma_{\text{layer}}
\end{bmatrix}}_{\text{each entry }\in\mathbb{R}^{C\times 1}}
=
\begin{bmatrix}
\dfrac{1}{L_X^{'}}\displaystyle\sum_{t}\mathbf X_{:,t} &
\sqrt{\dfrac{1}{L_X^{'}}\displaystyle\sum_{t}\bigl(\mathbf X_{:,t}-\mu_{\text{inst}}\bigr)^2+\varepsilon} \\[10pt]
\dfrac{1}{C'L_X^{'}}\displaystyle\sum_{c,t}\mathbf X_{c,t}\,\mathbf 1_{C'} &
\sqrt{\dfrac{1}{C'L_X^{'}}\displaystyle\sum_{c,t}\bigl(\mathbf X_{c,t}-\mu_{\text{layer}}\bigr)^2+\varepsilon}
\end{bmatrix},
\qquad
\varepsilon=10^{-3}.
\end{equation}
where $\varepsilon$ is a regularization term for numerical stability and $\mathbf 1_{C}$ is a unit vector. A 1-D Gating Network
$G(\cdot)=\sigma(\mathrm{Conv\!-\!ReLU\!-\!Conv})$
is used to compute per-channel weights $\mathbf g\!\in\![0,1]^{C\times1}$ from
$\mathbf C$, adaptively mixing statistics:
\begin{align}
  \mu = \mathbf g\odot\mu_{\text{inst}} + (1-\mathbf g)\odot\mu_{\text{layer}},\quad
  \sigma = \mathbf g\odot\sigma_{\text{inst}} + (1-\mathbf g)\odot\sigma_{\text{layer}}.
\end{align}
The content feature is then normalized by the adaptively gated mean $\mu$ and standard deviation $\sigma$ as: $\hat{\mathbf X}=(\mathbf X-\mu)/\sigma$.

\paragraph{Adaptive Affine Modulation.} The conditional feature was averaged by a global average pooling (GAP) as $\bar{\mathbf Y}=\tfrac1L\sum_t\mathbf Y_{:,t}$. This averaged feature is then passed to the Affine Network $\psi$, which consists of a multi-layer perceptron (MLP). The output is split into two affine coefficients $\gamma$ and $\beta$:
\begin{align}
[\gamma,\beta]=\psi(\bar{\mathbf Y}),\quad \mathbf H=(1+\gamma)\odot\hat{\mathbf X}+\beta.
\end{align}

\paragraph{Dynamic Masking.}
Channel importance scores
$s_c=\frac1L\sum_t|\mathbf Y_{c,t}|$ are computed to select the top-$k$ channels
($k=\lfloor\rho Y\rfloor$, $\rho=0.7$) among the modulated features.  Additionally, a binary mask $\mathbf M\!\in\!\{0,1\}^{C}$ is applied:
\begin{align}
  \tilde{\mathbf H}_{c,:}= \mathbf M_{c}\,\mathbf H_{c,:}.
\end{align}
Finally, the fused latent feature $\tilde{\mathbf H}$ is residually fused with the original prompt/text embeddings before being fed into a DiT decoder. Because DGF operates on arbitrary $(C,L)$ tensors, it handles four types of modality fusion in LoongX: EEG-PPG, fNIRS-Motion, neural-prompt, and neural-pooled-prompt.

\subsection{Conditional Diffusion}
The fused latent representation conditions a DiT backbone~\cite{peebles2023scalable} for image editing. The DiT model accepts the encoded input image $\mathbf I$ and fused latent feature $\tilde{\mathbf h}$ and outputs the edited image aligned with the semantic intention via fine-tuning.

Specifically, DiT predicts a velocity $\boldsymbol v_\theta(\mathbf I_t,\,t,\,\tilde{\mathbf h})$ that is used to iteratively refine the latent image in \(T\) uniform steps,
\begin{equation}
\mathbf I_{t-\frac{1}{T}}
= \mathbf I_t
\;-\;\frac{1}{T}\,
\boldsymbol v_\theta(\mathbf I_t,\,t,\,\tilde{\mathbf h}),
\qquad
\forall\,t\in\{1/T,2/T,\dots,1\}.
\label{eq:dit_update}
\end{equation}
At inference time we apply~\eqref{eq:dit_update} until \(t=0\), yielding the edited image \(\hat{\mathbf I}_0\).

\subsection{Pre-training and Finetuning}
We adopt a two-phase process:
1)~ neural signal encoders (EEG is the most important one) are pretrained on neuro-text corpora, compressing public data and L-Mind,
2)~ The full stack is optionally fine-tuned with paired original images and ground truth edited images.

\textbf{Pretraining.}
Signal encoders are pretrained to align with semantic embeddings using large-scale cognitive datasets~\citep{nieto2022thinking,ning2024fnirs} and L-Mind. CS3 encoders (EEG\,+\,PPG and fNIRS\,+\,Motion, respectively) are aligned to frozen text embeddings via symmetric NT-Xent loss:
\begin{align}
 \mathcal L_{\text{con}}
 &=\frac1{2M}\sum_{i=1}^{M}\!
   \Bigl[-\log\frac{e^{s_{ii}}}{\sum_j e^{s_{ij}}}
         -\log\frac{e^{s_{ii}}}{\sum_j e^{s_{ji}}}\Bigr].
 \label{eq:contrastive}
\end{align}
where $s_{ij}=(\mathbf{z}_i^{\top}\mathbf{q}_j)/\tau$, $\mathbf{z}_i$ and $\mathbf{q}_j$ are neural and text embeddings, and $M$ is the number of neural modalities. During pretraining, signal encoders are learned while text encoders stay frozen.

\textbf{Finetuning.}
Encoders and the DiT are finetuned jointly on L-Mind, mapping user neural patterns to editing target following a standard diffusion objective that minimizes the mean-squared velocity error. For an input image $\mathbf I _ 0$ and Gaussian noise $\boldsymbol\epsilon\!\sim\!\mathcal N(\mathbf 0,\mathbf I)$, where $\bar\alpha_t$ is the cumulated noise schedule:
\begin{equation}
\mathcal L_\mathrm{MSE}=\mathbb E_{t,\mathbf I_0,\boldsymbol\epsilon}
\bigl\|
\boldsymbol v_\theta(\mathbf I_t,t,\tilde{\mathbf h})
-
\underbrace{\bigl(\sqrt{\bar\alpha_t}\,\boldsymbol\epsilon-\sqrt{1-\bar\alpha_t}\,\mathbf I_0\bigr)}_{\boldsymbol v(\mathbf I_t,t,\boldsymbol\epsilon)}
\bigr\|_2^2.
\end{equation}

\section{Experiment}
\label{experiment}

% 这部分值得好好规划

\textbf{To answer the research questions (RQs) asked in Sec.~\ref{key:rqs}}, we conduct a comprehensive evaluation to validate the effectiveness of LoongX on the test set of L-Mind. This section first describes the experimental setup, evaluation metrics, and implementation details, and presents results of comprehensive quantitative evaluations, detailed breakdown analyses, and qualitative assessments.

\subsection{Experimental Setup}
\label{subsubsec:setup}

\paragraph{Implementation Details.}

All models are trained on eight NVIDIA H100 GPUs.  
Text prompts are embedded by T5-XXL~\cite{2020t5} and CLIP~\cite{radford2021learning}; neural signal streams are encoded by the proposed \textsc{CS3}.  
Unless stated otherwise, EEG montage (Fz, Fp2, O2, Pz, Cz) is sampled at 256\,Hz and down-sampled to 32\,Hz after band-pass filtering.  
Inference runs at 8 steps with classifier-free guidance $w=4$. We choose OminiControl~\citep{tan2024ominicontrol} as our baseline as it supports the text-conditioned image-editing based on DiTs. We also implement LoongX using only neural signals (EEG, fNIRS, PPG and Motion) and using both text prompts and neural signals. 
We load the pretrained weights from FLUX.1-dev\footnote{https://huggingface.co/black-forest-labs/FLUX.1-dev} and use low-rank approximation (LoRA) for fine-tuning (learning rate $1.0$, weight decay $0.01$).

\paragraph{Evaluation Metrics.}

We mainly use five metrics for quantitative assessment  following~\citep{zhang2023magicbrush}:

\begin{enumerate}[label=\arabic*)]
\item \textbf{L1 Distance (Mean Absolute Error)}: Calculating the average absolute difference between corresponding pixels in edited and ground truth tar images. 
\item \textbf{L2 Distance (Mean Squared Error)}: Computing the average squared difference between pixels. Penalizes large errors more heavily than L1.
\item \textbf{CLIP-I Score}: Evaluating semantic similarity between model-edited images and ground truth target images, which focuses on global semantics of editing results.
\item \textbf{DINO Score}: Assessing feature similarity between editing results and ground truth. Compared with CLIP, it is believed that DINO features capture fine-grained structural similarity that correlates with perceived preservation of identity, pose, and local geometry~\cite{zhou2024deformable}.

\item \textbf{CLIP-T Score}: Evaluating semantic similarity between image and textual prompts.
\end{enumerate}

\subsection{Reliability of Neural Signals}
\vspace{0.15cm}

\begin{theo}[\textbf{Answer \hyperref[key:rq1]{RQ1}} ]{thm:rq1}
\textbf{Answer \hyperref[key:rq1]{RQ1}}: Neural signals can serve as reliable indicators to drive image editing, outperforming text-instructed baselines on key metrics.
\end{theo}

\vspace{-0.2cm}

\begin{table}[!htbp]
  \centering
  \small
  \vspace{-0.5em}
  \caption{Comparison between baseline methods and two LoongX paradigms: (i) neural signals only and (ii) neural signals enhanced by speech. Mean $\pm$ 95\% confidence interval (CI) over three runs.}
  \vspace{-0.5em}
  \renewcommand{\arraystretch}{1.25}
  \setlength{\tabcolsep}{1mm}
  \begin{tabular}{lccccc}
    \toprule
    Methods & L1 ($\downarrow$) & L2 ($\downarrow$) & CLIP-I ($\uparrow$) & DINO ($\uparrow$) & CLIP-T ($\uparrow$) \\ \midrule[1.2pt]
    OminiControl (Text)        & \err{0.2632}{0.006} & \err{0.1161}{0.010} & \err{0.6558}{0.010} & \err{0.4636}{0.017} & \err{0.2549}{0.008} \\
    OminiControl (Speech)      & \err{0.2714}{0.006} & \err{0.1209}{0.008} & \err{0.6146}{0.009} & \err{0.3717}{0.013} & \err{0.2501}{0.008} \\ \midrule
    \rowcolor{gray!8}
    LoongX (Neural Signals)    & \berr{0.2509}{0.006} & \berr{0.1029}{0.009} & \berr{0.6605}{0.009} & \berr{0.4812}{0.015} & \err{0.2436}{0.008} \\
    \rowcolor{gray!8}
    LoongX (Signals+Speech)    & \err{0.2594}{0.006}  & \err{0.1080}{0.009}  & \err{0.6374}{0.009}  & \err{0.4205}{0.014}  & \berr{0.2588}{0.009} \\
    \bottomrule
  \end{tabular}
  \vspace{-1em}
  \label{tab:result_comparison}
\end{table}

As shown in Table~\ref{tab:result_comparison}, neural-signal-only LoongX outperforms the text-based OminiControl baseline in semantic discriminability (CLIP-I: $0.6605$ vs. $0.6558$) and robustness (DINO: $0.4812$ vs. $0.4636$), which demonstrates the potential of neural signals as a standalone modality that carries rich semantic information for image editing. The slightly higher L1 and L2 errors indicate better preservation of semantic fidelity over pixel-level accuracy. Combining speech cues with neural signals boosts semantic alignment, reaching the highest CLIP-T score of 0.2588 and demonstrating their joint effectiveness in capturing nuanced user intentions for hands-free image editing.

\begin{figure*}[t]
  \vspace{-0.8em}
  \centering
  \begin{minipage}{0.47\linewidth}
    \centering
    \includegraphics[width=\linewidth]{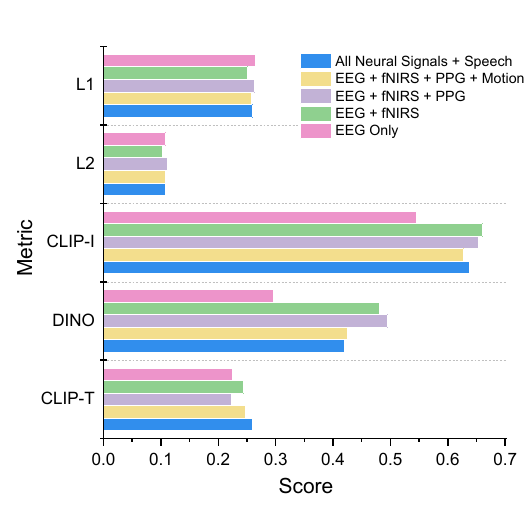}
    \vspace{-3em}
    \caption{Evaluation of different signal combinations on the proposed DGF module.}
    \label{fig:ablation_signals}
  \end{minipage}
  \hfill
  \begin{minipage}{0.48\linewidth}
    \centering
    \includegraphics[width=\linewidth]{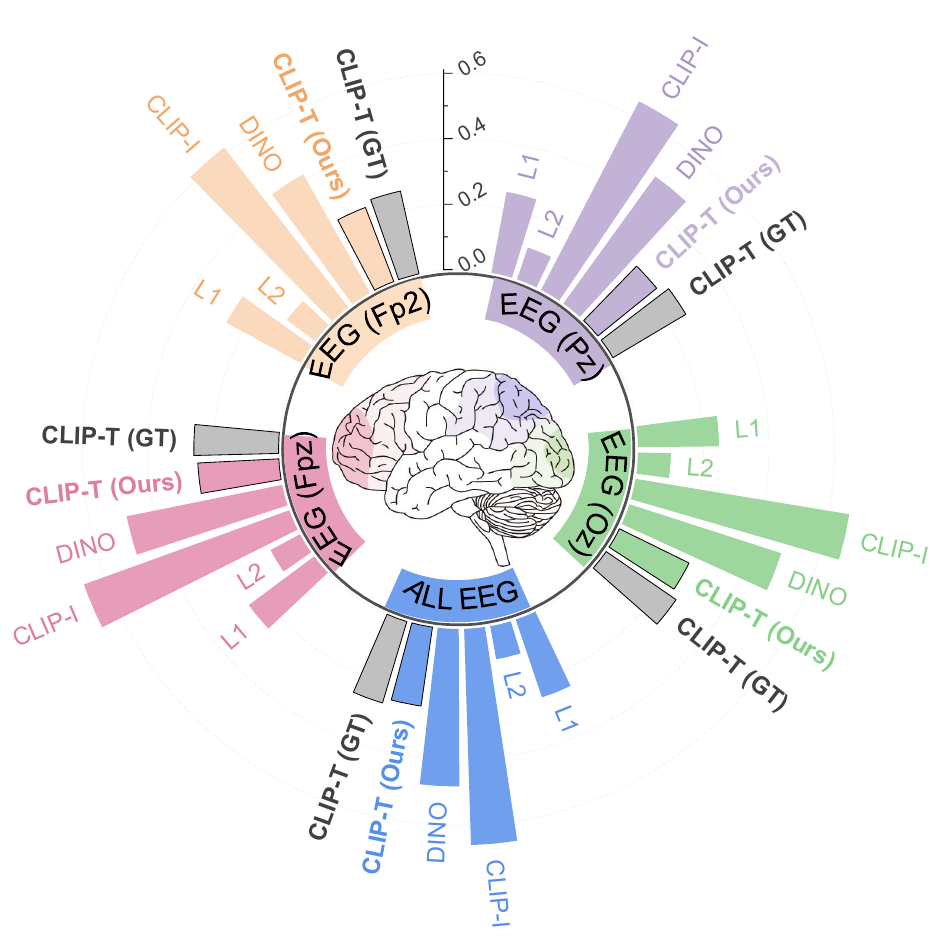}
    \vspace{-1.9em}
    \caption{Evaluation results on different brain region signals where LoongX is trained and tested on each respective EEG channel.}
    \label{fig:ablation_regions}
  \end{minipage}
  \vspace{-1.5em}
\end{figure*}

\subsection{Ablation Studies on Modality Contribution}
\vspace{0.15cm}
\begin{theo}[ ]{thm:rq2}
\textbf{Answer \hyperref[key:rq2]{RQ2}:} Different neural signal modalities contribute complementary strengths, enhancing discriminability, robustness, and semantic precision, respectively.
\end{theo}
\vspace{-0.2cm}

Modality contributions are compared in Fig.~\ref{fig:ablation_signals}. EEG signals alone enable basic high-level semantic editing, supported by the semantic discriminability of the extracted features (CLIP-I: $0.5457$). Integrating fNIRS significantly improves feature robustness (DINO: from $0.2963$ to $0.4811$), highlighting the complementary nature of hemodynamic responses in enhancing signal completeness and structural fidelity. Including PPG and Motion improves global physiological awareness and indicates sensitivity to subtle engagement patterns (e.g., heart rate and user movements) that express editing intent. They both contribute to the features' robustness and completeness to ensure stable CLIP-T score gains. 

We show the contribution of each individual EEG channel in Fig.~\ref{fig:ablation_regions}, where each channel corresponds to a specific scalp region as detailed in Table~\ref{tab:neuro_roles_all}. The occipital cortex channel (Oz), which is in charge of visual processing, emerges as dominant in global editing effect (CLIP-I: $0.6619$) and robustness (DINO: $0.4873$) to finer details, affirming its critical role in basic visual perception and processing tasks. Conversely, the frontopolar cortex (Fpz) provides superior semantic alignment (CLIP-T: $0.2481$), consistent with its association with more complex cognitive processes. Specifically, Fpz provides decision control and attention regulation compared with basic visual perception provided by Oz, which precisely confirms the discovery patterns in medical anatomy. This channel-specific analysis provides insights valuable for targeted applications or constrained hardware settings.

\subsection{Breakdown Analysis: Neural vs. Language-based Conditions}

\vspace{0.15cm}
\begin{theo}[\textbf{Answer \hyperref[key:rq3]{RQ3}} ]{thm:rq3}
\textbf{Answer \hyperref[key:rq3]{RQ3}:} Neural signals excel in low-level visual edits, while language excels in high-level semantics; combining both yields comprehensive and optimal control.
\end{theo}
\vspace{-0.2cm}

\begin{wrapfigure}{R}{0.5\textwidth}
    \centering
    \vspace{-2em} % 向下移动图像
    \includegraphics[width=0.6\textwidth]{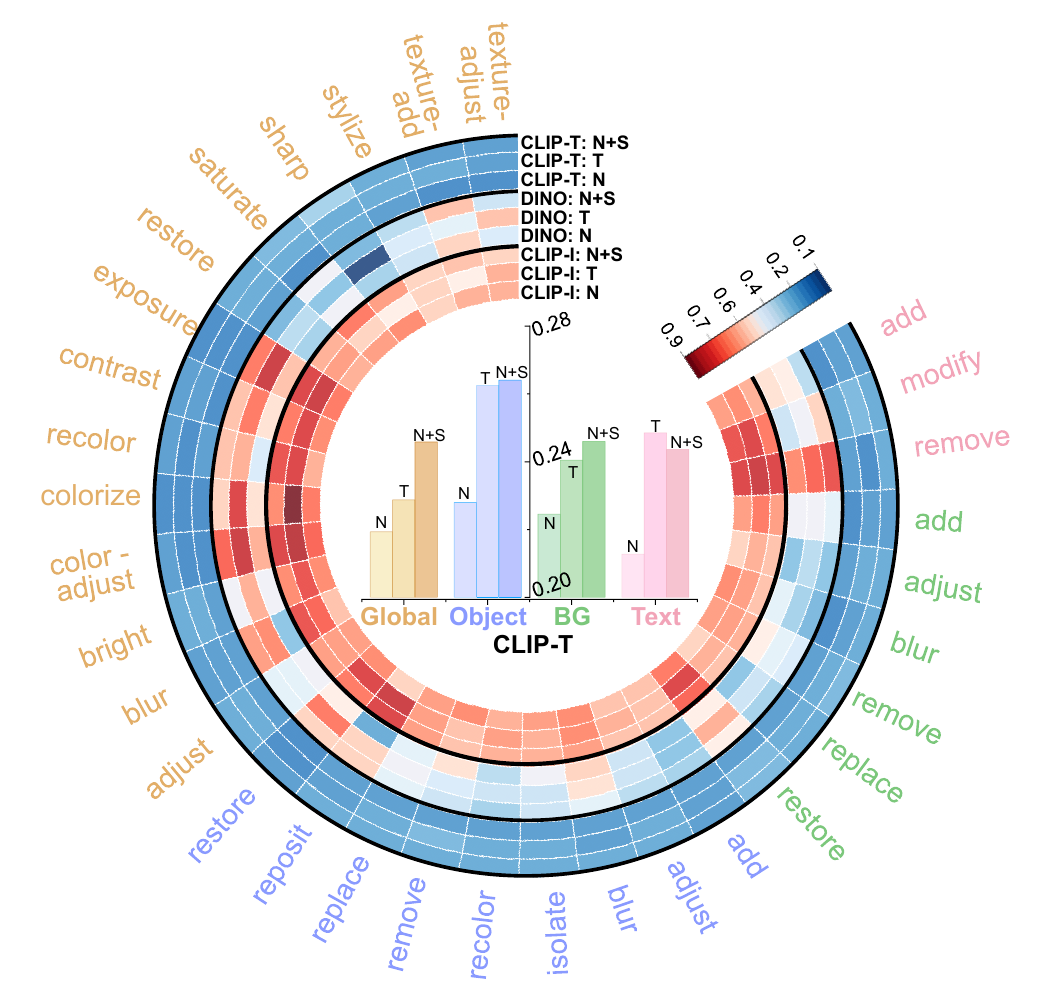} % 图像缩小
    \vspace{-2.5em} 
    \caption{Breakdown results of text and neural-driven image editing. BG: background.}
    \label{fig:breakdown}
    \vspace{-\baselineskip}
\end{wrapfigure}

The analysis of text and neural-driven image editing is shown in Fig.~\ref{fig:breakdown}. Using pure neural signals (N) is particularly effective for global texture editing, with higher CLIP-I scores highlighting their strong visual and structural consistency. 
Neural signals also outperform other modalities in several tasks like object removal and background blur, reflecting their strength in conveying intuitive intent, though they remain limited in handling complex semantics like text editing. Text instructions (T) are inherently stronger in high-level semantic tasks (e.g., image restoration), which indicates their advantage in describing instruction details. Combined neural and speech (N+S) signals achieve the highest semantic alignment (CLIP-T: $0.2588$), showcasing the superior effectiveness of hybrid conditioning in capturing complex user intentions. Overall, neural signals are more effective for low-level visual edits, while neural and text-based approaches each provide complementary advantages.

\subsection{Ablation Studies on Model Architecture}

% \vspace{0.15cm}
% \begin{theo}[\textbf{Answer \hyperref[key:rq3]{RQ3}} ]{thm:fd3}
% \textbf{Finding \hyperref[key:fd1]{1}:} Each component in the LoongX architecture contributes uniquely, and their composition (especially with pretraining) maximizes the performance potential, leading to improved editing accuracy across metrics.
% \end{theo}

\vspace{-0.2cm}

\textbf{Each component in the LoongX architecture contributes uniquely, and their composition (especially with pretraining) maximizes the performance potential, leading to improved editing accuracy across metrics.}
The ablation study in Table~\ref{tab:ablation} is conducted on the fused neural signals and speech to evaluate the impact of each proposed module. It is found that CS3 encoder enhances feature completeness and smoothness, leading to a 5\% reduction in L2 error. The DGF module improves semantic alignment with textual instructions, yielding a 3.5\% increase in CLIP-T. Supplemented by pre-training, LoongX reaches optimal performance, indicating the important role of robust and representation learning and multimodal alignment in maximizing editing performance.

\begin{table}[ht]
  \centering
  \small
  \vspace{-0.3em}
  \caption{Ablation studies on the architecture of LoongX. Mean $\pm$ 95\% CI over three runs.}
  \vspace{-0.5em}
  \renewcommand{\arraystretch}{1.25}
  \setlength{\tabcolsep}{1.25mm}
  \begin{tabular}{ccc ccccc}
    \toprule
    Pretrain & CS3 & DGF &
    L1 ($\downarrow$) & L2 ($\downarrow$) & CLIP-I ($\uparrow$) & DINO ($\uparrow$) & CLIP-T ($\uparrow$) \\
    \midrule[1.2pt]
      &  &  & \err{0.2645}{0.005} & \err{0.1099}{0.003} & \err{0.5966}{0.009} & \err{0.3948}{0.011} & \err{0.2584}{0.010} \\
      & \checkmark &  & \berr{0.2567}{0.006} & \berr{0.1047}{0.004} & \berr{0.6408}{0.010} & \berr{0.4588}{0.012} & \err{0.2248}{0.007} \\
      &  & \checkmark & \err{0.2629}{0.005} & \err{0.1106}{0.003} & \err{0.6025}{0.009} & \err{0.3992}{0.011} & \berr{0.2620}{0.007} \\
      & \checkmark & \checkmark & \err{0.2648}{0.006} & \err{0.1124}{0.004} & \err{0.6319}{0.009} & \err{0.4162}{0.012} & \err{0.2534}{0.008} \\
    \rowcolor{gray!8}
    \checkmark & \checkmark & \checkmark & \berr{0.2594}{0.006} & \berr{0.1080}{0.004} & \berr{0.6374}{0.009} & \err{0.4205}{0.012} & \berr{0.2588}{0.009} \\
    \bottomrule
  \end{tabular}
  \vspace{-1em}
  \label{tab:ablation}
\end{table}

\subsection{Qualitative Analysis and Discussion}

% \vspace{0.15cm}
% \begin{theo}[\textbf{Answer \hyperref[key:rq3]{RQ3}} ]{thm:fd2}
% \textbf{Finding \hyperref[key:fd2]{2}:} LoongX demonstrates image editing capabilities from neural signals, though performance may vary slightly with complex or ambiguous intentions.
% \end{theo}
% \vspace{-0.2cm}

\begin{figure*}[t]
	\centering
	\includegraphics[trim=0cm 0.2cm 0 0, width=1.0\textwidth]{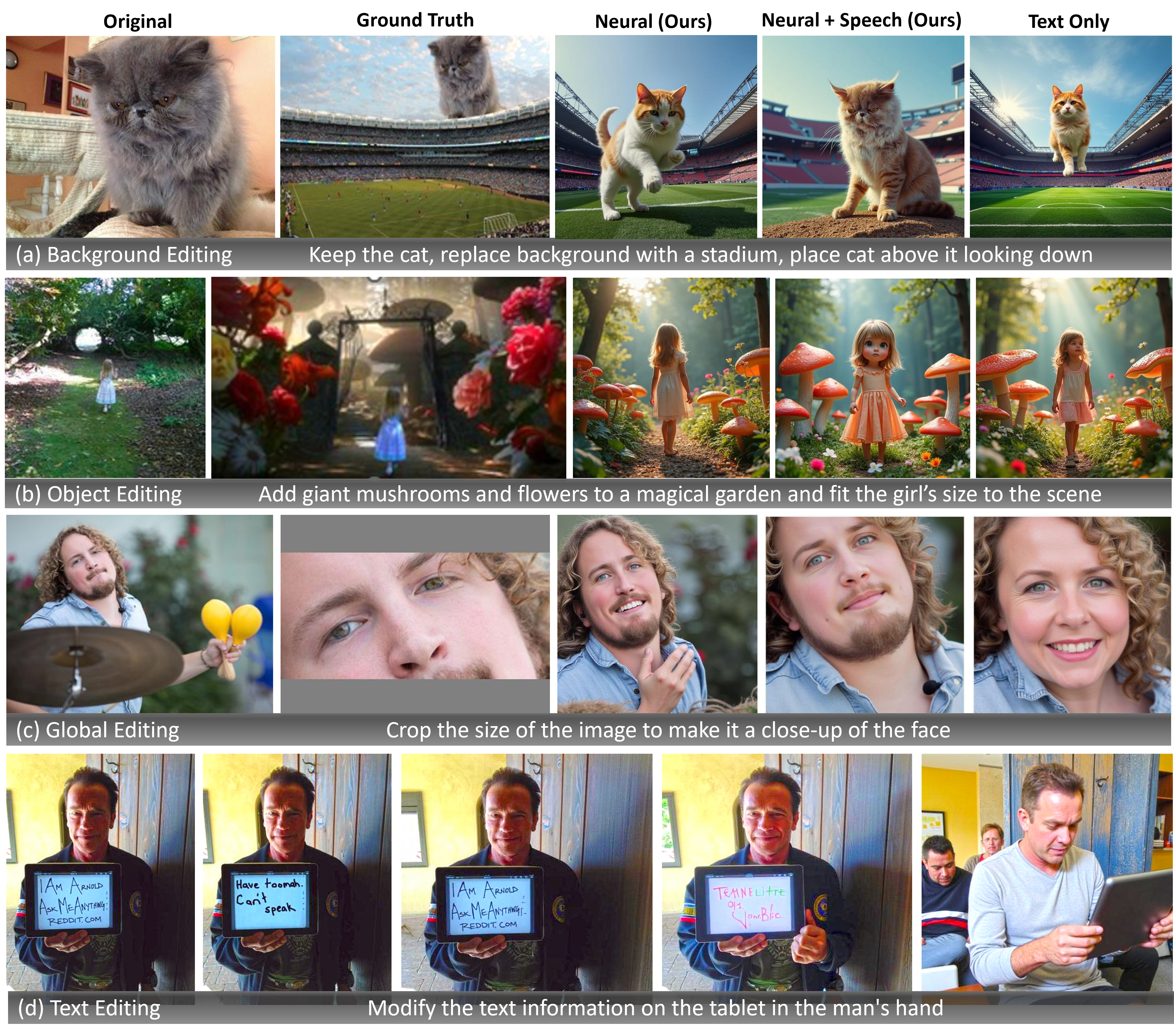}
    \vspace{-2em}
  \caption{Qualitative evaluation comparing the text prompt-based method and our neural-driven methods for four editing categories: (a) background, (b) object, (c) global, and (d) text editing.}
	\label{fig:cases}
    \vspace{-1em}
\end{figure*}

\textbf{Qualitative examples confirm LoongX's intuitive editing capabilities, though performance may vary slightly in scenarios with complex or ambiguous intentions.}
Qualitative results presented in Fig.~\ref{fig:cases} demonstrate that neural-driven LoongX can successfully achieve various visual and structural modifications, such as background replacement and global adjustments. 
However, the fused neural-language method better captures nuanced instructions involving abstract semantics (e.g., ``modify the text information" in Fig.~\ref{fig:cases}(d)). 

\paragraph{Limitations and future work.}
It is noted that while the neural signals and combined methods perform better in multiple tasks such as object manipulation (e.g., letting the cat look down in Fig.~\ref{fig:cases}(a)), the text-based method handles spatial manipulation more effectively (e.g., ``place the cat above" in Fig.~\ref{fig:cases}(a)).
Despite significant advancements with multimodal fusion, entity consistency (e.g., the style of the little girl in Fig.~\ref{fig:cases}(b)) remains a challenge, which is limited by the capabilities of backbone image editing models at the time this work was mainly completed. Moreover, highly abstract or ambiguous instructions occasionally still pose challenges (e.g., ``winged white animal in Fig.~\ref{fig:sup_case2}(f)"). 
Several more failure cases are shown in Fig.~\ref{fig:sup_case5}, indicating areas where further refinement in entity interpretation and neural-data-based disambiguation remains necessary.

Overall, the experiments validate LoongX's efficacy as a robust, intuitive interface leveraging neural signals for image editing. Multimodal integration, particularly neural and linguistic fusion, emerges as essential for capturing comprehensive user intent. These findings advocate for future research focusing on improving semantic interpretation fidelity from neural signals and exploring adaptive methodologies for further enhancing accessibility and precision in cognitive-driven creative technologies. Incremental fine-tuning for new unseen users is also worth exploration in practice.

% \paragraph{Discussion and Future Directions.}
% Overall, the experiments confirm that (i) neural signals alone provide competitive control, (ii) fNIRS and frontal EEG chiefly benefit semantic alignment, and (iii) \textsc{CS3}+\,\textsc{DGF}+pre-train is vital for stable convergence.  
% Overall, the experiments validate LoongX's efficacy as a robust, intuitive interface leveraging neural signals for image editing. Multimodal integration, particularly neural and linguistic fusion, emerges as essential for capturing comprehensive user intent. These findings advocate for future research focusing on improving semantic interpretation fidelity from neural signals and exploring adaptive methodologies for further enhancing accessibility and precision in cognitive-driven creative technologies. Incremental fine-tuning for new unseen users is also worth exploration in practice.
\section{Conclusion}
\label{conclusion}

We presented LoongX, a novel framework for hands-free image editing by conditioning diffusion models on multimodal neural signals, achieving performance comparable to or superior to traditional text-driven baselines.
Looking ahead, the portability of our wireless setup opens exciting possibilities for real-world applications in immersive virtual environments. Future directions include integrating LoongX with VR/XR to support intuitive cognitive interaction and aligning neural representations with emerging world models~\cite{mao2025yume,ni2025recondreamer} to project human intention into an interactive synthetic world, enabling mind-driven control in virtual realities.

\begin{ack}

We sincerely thank Prof. Tat-Seng Chua (National University of Singapore) for his valuable comments and suggestions during the drafting of this work. We also thank Dr. Zhiwei Tang (Alibaba) for his constructive guidance during the rebuttal. This work is sponsored by the NUS Startup Grant (Presidential Young Professorship), Singapore MOE Tier-1 Grant, ByteDance Grant, NUS ARTIC Grant, Apple Grant, Alibaba Grant, and Google Grant for TPU usage, and also supported by the STI2030-Major Projects (No. 2021ZD0200401), Zhejiang Province Key R\&D Programs (No. 2024C03001, 2024C03007, 2024C01031, 2025C01137, 2025C02165), Zhejiang Province High-Level Talent Special Support Plan (No. 2022R52042), and the Fundamental Research Funds for the Central Universities (No. 2025ZFJH01).

\end{ack}
%%%%%%%%%%%%%%%%%%%%%%

% \section*{References}

{
    %\small
    %\bibliographystyle{bibstyle}
    \bibliographystyle{unsrt}
    \bibliography{references}
}

%  Any choice of citation style is acceptable as long as you are consistent. It is permissible to reduce the font size to \verb+small+ (9 point) when listing the references.

% \medskip

% {
% \small
% [1] Alexander, J.A.\ \& Mozer, M.C.\ (1995) Template-based algorithms for
% connectionist rule extraction. In G.\ Tesauro, D.S.\ Touretzky and T.K.\ Leen
% (eds.), {\it Advances in Neural Information Processing Systems 7},
% pp.\ 609--616. Cambridge, MA: MIT Press.

% [2] Bower, J.M.\ \& Beeman, D.\ (1995) {\it The Book of GENESIS: Exploring
%   Realistic Neural Models with the General Neural Simulation System. New York:
% TELOS/Springer--Verlag.

% [3] Hasselmo, M.E., Schnell, E.\ \& Barkai, E.\ (1995) Dynamics of learning and
% recall at excitatory recurrent synapses and cholinergic modulation in rat
% hippocampal region CA3. {\it Journal of Neuroscience} {\bf 15}(7):5249-5262.
% }

%%%%%%%%%%%%%%%%%%%%%%%%%%%%%%%%%%%%%%%%%%%%%%%%%%%%%%%%%%%%

\clearpage

\appendix

\section{Supplementary Key Information}
\label{sec:appendix}

\subsection{Preliminary: Analysis of Brain Region Function and Mechanism}

\begin{figure}[H]
    \centering
    \includegraphics[trim=0cm 4cm 0cm 4.5cm, clip, width=0.98\linewidth]{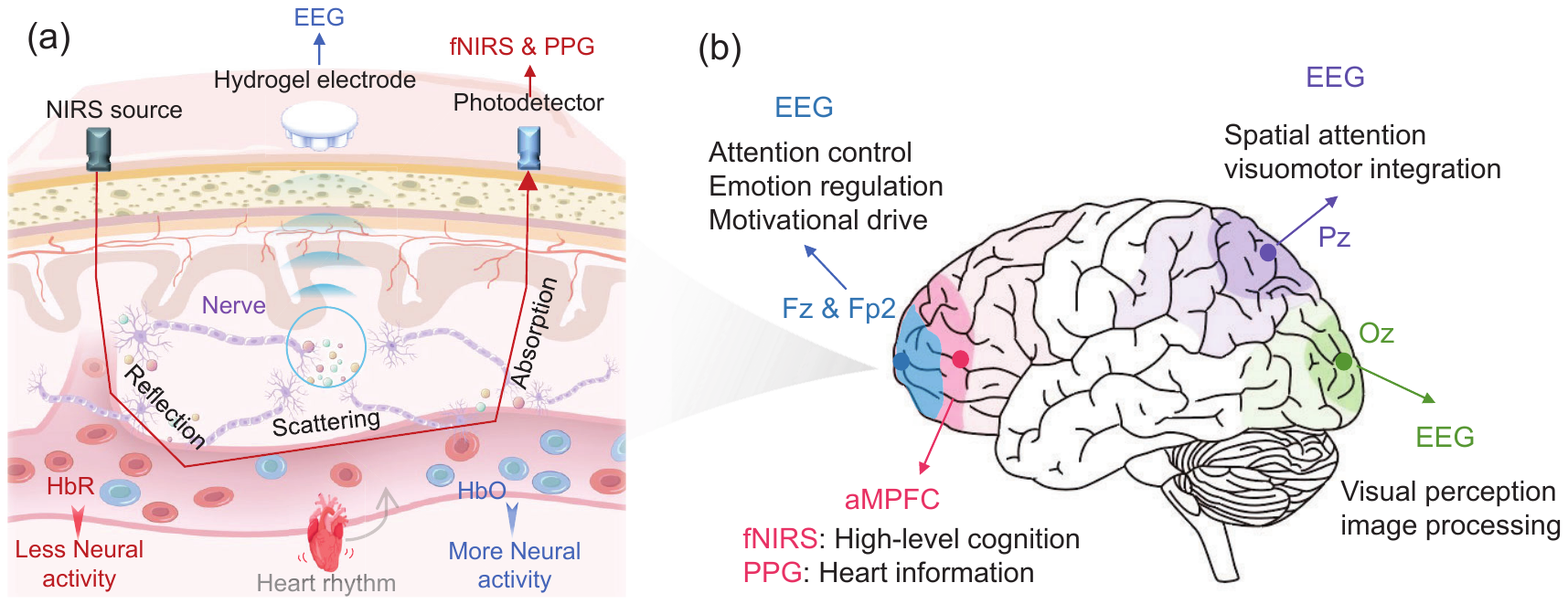}
    \vspace{-0.3cm}
    \caption{(a) Signal acquisition mechanism. (b) Cognitive function of different brain regions.}
    \vspace{-0.3cm}
    \label{fig:collectMach}
\end{figure}

We employ a noninvasive multimodal sensing system that synchronously records neural, hemodynamic, and peripheral vascular signals, including EEG, fNIRS, and PPG (Fig.~\ref{fig:collectMach}(a)). This setup enables comprehensive monitoring of brain activity during human–computer interaction.
fNIRS employs near-infrared light in the 690–930 nm range to penetrate scalp and skull tissues. Neural activation leads to increased local cerebral blood flow and metabolic demand, resulting in a rise in oxygenated hemoglobin (HbO) and a reduction in deoxygenated hemoglobin (HbR). These changes cause detectable variations in light absorption at the detector. fNIRS thus provides second-level sensitivity to the slow hemodynamic responses associated with cortical processing.
EEG is recorded via hydrogel-based electrodes placed over the scalp, capturing millisecond-scale voltage fluctuations resulting from synchronized postsynaptic potentials in cortical pyramidal neurons. This modality offers high temporal precision for observing rapid fluctuations in cortical excitability and sensorimotor responses.
PPG detects volumetric changes in blood flow via near-infrared light, enabling continuous measurement of pulse rate and vascular compliance. The sensor is co-located with fNIRS optodes, sharing similar optical pathways but tuned for peripheral cardiovascular features.

This multimodal approach provides a synergistic view of neural function, combining the high temporal resolution of EEG with the metabolic and vascular insights from fNIRS and PPG. Such cross-modal sensing is crucial for modeling the perception-decision-action cycle in neuroadaptive systems.

\begin{table*}[htbp]
\centering
\small
\caption{Functional Roles of Multimodal Neural Signals in Hands-Free Image Editing. }
\begin{tabular}{p{0.16\linewidth} p{0.17\linewidth} p{0.22\linewidth} p{0.34\linewidth}} 
\midrule
% \textbf{Signal/Channel} & \textbf{Cortical/Physiological Region} & \textbf{Primary Function} & \textbf{Role in Image Editing} \\
\textbf{Signal (Channel)} & \textbf{Cortical Region} & \textbf{Primary Function} & \textbf{Roles in Image Editing} \\
\midrule
EEG Ch 0 (Pz) & Parietal cortex & Spatial attention, visuomotor integration & Focuses on specific image areas, targets object localization~\citep{Frontoparietal1}  \\
\midrule
EEG Ch 1 (Fp2) & Prefrontal cortex & Emotion regulation, motivational drive & Generates and regulates intentional editing actions~\citep{Sengupta2024}  \\
\midrule
EEG Ch 2 (Fpz) & Frontopolar cortex & Attention control, task initiation & Triggers editing intention, starts/stops editing operations~\citep{Friedman2022, Kim2021}  \\
\midrule
EEG Ch 3 (Oz) & Occipital cortex & Visual perception, image processing & Perceives visual changes, evaluates whether edits meet expectations~\citep{Ozkara2019ERP} \\
\midrule
fNIRS (aMPFC) & Anterior medial prefrontal cortex & High-level cognition, motivation, emotional valence & Indicates editing intent intensity, emotional confidence, and mental workload~\citep{AGRAWAL2025299} \\
\midrule
PPG (aMPFC) & Cardiovascular or autonomic system & Heart rate variability, arousal, stress & Monitors cognitive stress or emotional arousal during editing ~\citep{Arsalan2021,Tazarv2021}\\
\midrule
\end{tabular}
\label{tab:neuro_roles_all}
\end{table*}

To enable hands-free image editing, we integrate multimodal neural and physiological signals to decode user intent and cognitive state in real time. The functional roles of multimodal neural signals are provided in Table.\ref{tab:neuro_roles_all}. Specifically, midline parietal EEG (Pz) reflects spatial attention and visuomotor integration, supporting the allocation of attention to target areas and coordination of motor plans during editing tasks~\citep{Frontoparietal1}. Right prefrontal EEG (Fp2) is linked to emotional regulation and motivational drive, aligning with findings from frontal alpha asymmetry studies~\citep{Sengupta2024}. Frontopolar EEG (Fz/Fpz) tracks attention control and task initiation, facilitating the onset and modulation of editing operations~\citep{Friedman2022, Kim2021}. Occipital EEG (Oz) encodes visual perception and image processing load, enabling evaluation of visual changes and edit quality~\citep{Ozkara2019ERP}. fNIRS over anterior medial prefrontal cortex (aMPFC) measures cognitive load, motivation, and emotional valence through hemodynamic activity~\citep{AGRAWAL2025299}. PPG signals from the same region capture heart rate variability and autonomic arousal, which reflect user stress and engagement~\citep{Arsalan2021,Tazarv2021}.
This multimodal mapping (Fig.~\ref{fig:collectMach}(b)) enables the system to continuously adapt to users’ cognitive focus, affective state, and mental workload, thereby supporting a more responsive and intuitive editing experience.

\subsection{Preliminary Analysis: Editing Type Classification Experiment}
\label{subsec:classification_experiment}

LoongX is proposed to address three key research questions:

\begin{enumerate}
    \item Can neural signals serve as reliable conditions for image editing? (Does it really work?)
    \item What kind of information is conveyed by multimodal neural signals? (What do they actually contribute to image editing?)
    \item How do neural-based and language-based conditions differ in image editing? Can we combine their strengths to enable hands-free editing more effectively?
\end{enumerate}

In response to these problems, we conduct a premise exploration based on a classification experiment and design the LoongX model based on the findings. Finally, we present a modular architecture comprising unified multimodal encoding, dynamic multimodal data fusion, and diffusion-based conditional generation. Based on these, LoongX can perform robust hands-free image editing by translating user neural states into structured conditions for a diffusion model. 

To examine whether neural signals can reliably encode semantic conditions for image editing, we perform an exploratory classification experiment where the task is to predict editing types from neural signals or text. We use the 22,691 training instances and 1,200 test instances in L-Mind. As each editing instance can involve multiple editing types, we implement a simple multilayer perception (MLP) with three nonlinear-activated linear layers and conduct a multi-label classification experiment that recognizes all involved editing type labels for an instance, via text embeddings or neural signals, as a condition. We compare models trained with random noise, only text prompts, EEG, fNIRS, PPG, combinations of multimodal neural signals, and fusion of both text and signals.

\begin{figure*}[t]
	\centering
	\includegraphics[trim=2.4cm 0cm 0.8cm 0.5cm,clip, width=1.0\textwidth]{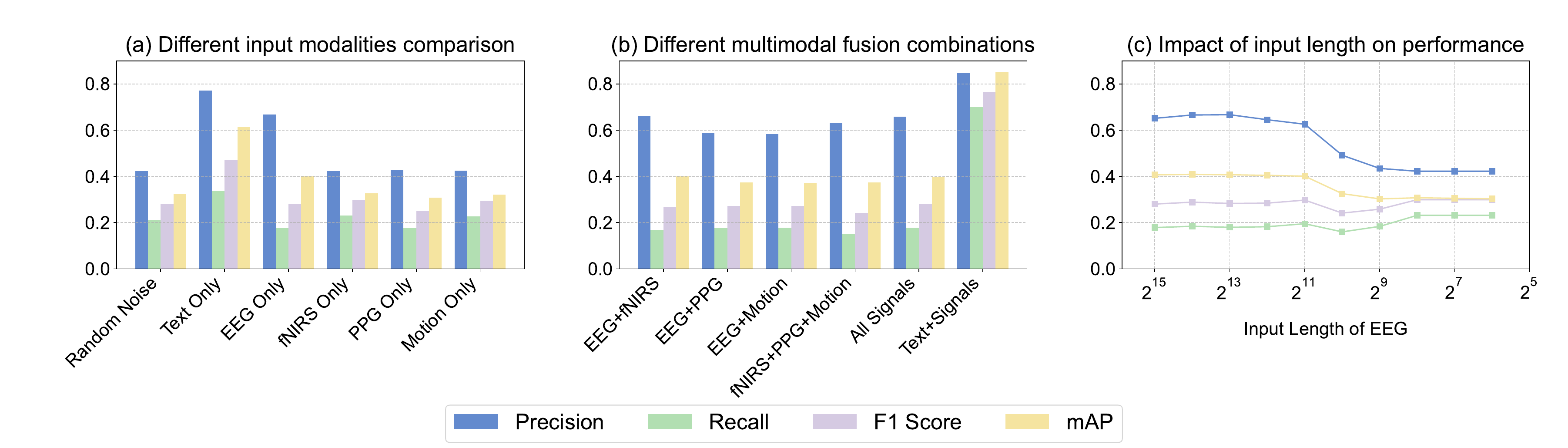}
 \caption{Multi-label classification result under different settings: (a) different single input modalities; (b) different modality fusion combinations; (c) different EEG input sequence lengths.}
	\label{fig:classification}
\vspace{-0.5cm}
\end{figure*}

As shown in Fig.~\ref{fig:classification}, the classification results validate the informativeness and complementarity of neural signals for image editing conditions. Fig.~\ref{fig:classification}(a) shows that using neural signals, especially the EEG signal itself, can achieve significantly better classification performance than random noise (as seen from over 7\% mAP improvement). fNIRS contributes to recall performance gain compared with random noise since it provides more complete and robust information, which is more important for recall. As image editing requires discriminant semantic features, it shows that neural networks can still effectively recognize the editing types (over 60\% precision) based on EEG signals and can even achieve comparable performance with text prompts in some cases. Results can also initially respond to question 2. As precision depends on the discriminability of features, EEG contributes more discriminability with less noise.  For recall, robustness and completeness contribute more. Therefore, fNIRS becomes more stable with less fluctuation. Though PPG and motion do not affect classification performance significantly as their volatility is small, they are expected to provide richer background information to stabilize performance. Now, what will the performance change if we combine these signals?

Fig.~\ref{fig:classification}(b) shows that binding all neural signals (via simple concat and MLP) achieves the best performance, and fusing only EEG and fNIRS can achieve comparable performance with only a recall reduction. Moreover, though neural signals are not more discriminant and robust than text embeddings, fusing both can achieve even better performance than text only. That is, the two types of data can complement the missing key information to achieve better category recognition in image editing tasks, which shows higher performance as a more powerful condition for further generative models.

As the neural signals all have different shapes, unifying the input sequence into a unified length is necessary. Simple padding and truncating to a fixed length is one of the most reliable methods for preserving most information in raw signals. We also investigate the influence of input sequence length for EEG as a neural signal representative. Results in Fig.~\ref{fig:classification}(c) show that truncating and padding EEG signal sequences to a unified 8,192 can achieve the best performance. While the longer sequence ensures a more reliable performance, the computation cost will become a burden as the sequence length increases. A trade-off method is needed to encode the most valid information in signals while not bringing unbearable computational costs. Therefore, we design the Cross-Scale State Space (CS3) model to ensure the best trade-off between performance and computational cost.

\subsection{More Limitations Discussion}
\label{subsec:limitation}

While LoongX demonstrates strong performance in neural-driven image editing, several limitations remain. First, the current dataset was collected from a relatively homogeneous group of 12 healthy young adults. Although the model performs well within this cohort, generalization to broader populations (e.g., different age groups or individuals with neurological conditions) is not yet fully validated. Moreover, the neural signals were acquired using low-density EEG and fNIRS systems, which, despite their practicality and portability, offer limited spatial resolution compared to high-density or invasive setups. This constraint may affect the system's ability to capture fine-grained neural representations.

The robustness of LoongX under varying data distributions and noisy conditions has not been systematically explored. Although some resilience is expected from the multimodal design and the DUAF fusion strategy, comprehensive stress testing against motion artifacts, sensor dropout, or environmental interference is a necessary next step for real-world deployment. Furthermore, while the BCI system we use is designed to work across participants, it may benefit from test-time adaptation or few-shot user-specific finetuning to account for individual variability in neural signatures, which can differ significantly across users.

Finally, the interpretability of the learned neural representations remains limited. While the CS3 encoder effectively distills signal patterns for downstream editing, it is not yet clear how these latent features relate to interpretable cognitive states or intentions. Improving the transparency and explanatory power of the system will be essential for broader acceptance and responsible deployment, especially in domains involving sensitive user data.

\subsection{Broader Societal Impacts}

The proposed neural-driven image editing technique has the potential for significant positive societal impact by improving accessibility for individuals with motor or communication impairments, enabling more inclusive creative workflows, and lowering the barrier to content creation. By enabling hands-free and intuitive interaction with generative models, this technology could foster greater participation in digital art, design, and communication, particularly for users with physical limitations.

However, as with other generative and brain-computer interface technologies, there are potential negative societal impacts. Malicious or unintended uses may include the creation of manipulated or deceptive media, privacy violations stemming from the misuse of neurophysiological data, or unauthorized surveillance. There are also concerns regarding fairness, as disparities in device availability or neural signal quality across user populations could exacerbate existing inequities.

To mitigate these risks, we recommend the implementation of safeguards such as user authentication, audit trails for sensitive editing actions, and transparent communication regarding data usage and model limitations. Responsible deployment should include ongoing monitoring and mechanisms to address feedback and misuse.

\subsection{Safety and Ethics}
All participants signed consent forms before the experiment, which was approved by the Institutional Review Board's Ethics Committee for Human Research Protections at Zhejiang University (Protocol ID: No. 067 (2019)).

Recognizing the potential for misuse of proposed models and multimodal neural datasets, we are committed to responsible release practices. We will require users requesting access to the models or data to undergo an application and review process, ensuring alignment with ethical guidelines and legitimate research or clinical objectives. Usage agreements will prohibit malicious activities, and access may be revoked in the event of violations.

For dataset release, we apply stringent filtering to remove personally identifiable, sensitive, or potentially harmful content. For model access, we will implement usage restrictions, safety filters, and ongoing monitoring to prevent abuse. Additionally, clear documentation outlining acceptable use cases, known limitations, and recommended best practices will be provided to all users.

We will continuously evaluate the impact and usage of the released resources and remain open to community feedback to improve the effectiveness of these safeguards over time.

\section{Supplementary Literature Review}

\subsection{Brain Computer Interface}

BCI is an emerging technology that establishes a direct communication pathway between the brain and external devices by interpreting neural signals. BCIs have broad applications in healthcare, communication, gaming, and assistive technologies~\citep{Edelman2025Noninvasive, Eldawlatly2024GenerativeBCI}. The primary goal is to help individuals, especially those with motor impairments, interact with digital systems without the constraints of traditional input methods.

Non-invasive BCIs, such as EEG and fNIRS, are widely studied for their safety and real-time signal acquisition capabilities~\citep{Nazeer2025fNIRS-EEG}. EEG captures the brain's electrical activity and is particularly effective in detecting cognitive states such as attention and intention. While fMRI provides high-resolution whole-brain imaging, its use is constrained by high cost and limited mobility. In contrast, fNIRS measures hemodynamic responses and offers a portable, cost-effective solution for monitoring brain activity in real-world environments (as illustrated in Fig.~\ref{fig:collect} and Fig.~\ref{fig:collectMach}).

With the rapid development of artificial intelligence, especially deep learning, BCI systems have seen significant improvements in decoding accuracy and robustness~\citep{lv2020advanced, zhang2018internet}. Advanced models enable tasks such as mental spelling, prosthetic control, and gaming interfaces. Recent trends include integrating BCIs with generative models such as diffusion networks to synthesize visual content directly from brain activity, as demonstrated in work applying stable diffusion to fMRI decoding~\citep{Takagi2023DiffusionBCI}. The integration of EEG and fNIRS holds great promise for improving the efficiency of human–AI interaction without compromising the portability that makes non-invasive systems suitable for real-world applications.

%In particular, the enhancement of BCI hardware and the accumulation of large-scale neurophysiological data, such as EEG and fNIRS, are vital for our research. EEG detects electrical activity in the brain and is effective for capturing cognitive states like attention and motor imagery. On the other hand, fNIRS, which measures hemodynamic responses, is particularly useful for tracking localized brain activity and cortical activation. The integration of these signals has promising potential to improve human-AI interaction efficiency.

\subsection{Brain-supervised Generation}

Advances in machine learning have significantly enhanced BCI decoding accuracy and robustness, opening new possibilities for brain-supervised generative methods that transform neural signals into visual content~\citep{Palazzo_2017_ICCV,chen2024mind}.

Several recent methods integrate neurophysiological data (e.g., fMRI, EEG, or fNIRS) with generative models~\citep{Takagi_2023_CVPR, bai2024dreamdiffusion, tripathy2021decoding}. For instance, CMVDM aligns fMRI features with semantics for image synthesis~\citep{zeng2024controllable}, and the MindEye series further lifts the resolution of generated images from decoded fMRI~\citep{Scotti_2023_NeurIPS,Scotti_2024_ICML}. 
DreamConnect translates brain signals into images based on fMRI, which is less accessible for everyday interaction~\cite{sun2025connecting}.
DreamDiffusion produces images from EEG via temporal masked modeling~\citep{bai2024dreamdiffusion}. EEG2Video extends the idea to dynamic video content~\citep{Liu_2024_NeurIPS}. While Davis~\textit{et al.}~\citep{Davis_2022_CVPR} initially explore brain-guided semantic image editing using a generative adversarial network, this work is limited to facial images and EEG signals. Moreover, Adamic~\textit{et al.}~\citep{adamic2024progress} reconstructs visual images from brain activity measured by fNIRS.

Unlike previous studies, our data were collected via a wireless BCI system, as shown in Fig.~\ref{fig:collect}, from participants performing instruction-based image editing tasks. Compared with fMRI-based methods, this multimodal setup is portable and suited to daily use, which can support greater portability and broader real-world applicability. To the best of our knowledge, our work is the first to comprehensively utilize the full information of \emph{EEG}, \emph{fNIRS}, \emph{PPG}, and head-motion signals for image editing. We specifically delve deeper into strategies of extracting neural features and fusing multimodal data, optimizing their integration to serve image editing needs.  

\subsection{Instruction-based Image Editing}

Instruction-based editing tasks include adjusting color, contrast, and brightness; retouching objects or backgrounds; and applying filters or artistic effects. This progress ranges from global editing (e.g., style transfer~\citep{brooks2023instructpix2pix}) to region-based image editing~\citep{zhao2024ultraedit}, which serves a wide range of applications in advertising, entertainment, and social media~\citep{zhao2024ultraedit}. 

The surge in multimodal data has fostered the development of large models capable of complex creative tasks~\citep{guo2025deepseek,bai2025qwen2}. Recent models such as GPT-4o~\citep{gpt4} and Gemini~\citep{team2023gemini} have evolved from basic data analysis to advanced image editing agents. These agents leverage generative methods to interpret user instructions for precise image manipulation.

Current instruction-based image editing agents integrate multimodal inputs, including text, images, and videos, to accurately identify and apply visual edits~\cite{shuai2024survey}. Leveraging learned multimodal representations, these agents interpret instructions from input, localize relevant regions, and perform targeted modifications~\cite{zhang2025context, tan2024ominicontrol, tan2025ominicontrol2}. Recent approaches, such as InstructPix2Pix~\citep{brooks2023instructpix2pix}, UltraEdit~\citep{zhao2024ultraedit}, MagicBrush~\citep{zhang2023magicbrush}, MIGE~\citep{tian2025mige}, and ACE~\citep{mao2025ace++} improve region-specific edits guided by natural language prompts. It is noticed that Jiang \textit{et al.}~\citep{jiang2021talk} explore speech-driven image editing, highlighting the feasibility of hands-free interaction but still limited by linguistic expressiveness in recorded speech.

Despite these advancements, achieving efficient, delicate, prompt-free image editing remains challenging. Our work addresses this gap, exploring neural-signal-driven editing agents to decode cognitive intent directly for image manipulation, significantly enhancing accessibility and interaction efficiency.

\subsection{Future Research Prospects}

Recent studies have expanded the frontiers of multimodal representation learning and diffusion-based generation, which benifits the future improvement of this work~\cite{li2025hyfacial,ni2025wonderturbo,xin2025resurrect,liu2025topolidm,zhao2024dynamic2,zeng2025FSDrive,zhang2025exploit}. For example, Zhou et al.~\cite{zhou2025hademif}, Xu et al.~\cite{xu2025causal} and Liu et al.~\cite{liu2025gated} extended causal and gated frameworks toward robust multimodal understanding. Li et al.~\cite{li2025multi} further introduced multi-objective unlearning, and Zeng et al.~\cite{zeng2025bridging} advanced precise text editing for controllable LLM adaptation. These developments collectively reflect a broader shift toward interpretable, data-efficient, and causally grounded multimodal methods, which contribute to future neural-driven AI systems~\cite{zhou2025valuing,zhao2024dynamic,zeng2025janusvln,10800533}.

Complementary progress has also been made in visual information processing, multimodal perception and generative modeling~\cite{zhao2024balf,zhao2025stitch,wang2025target,zhou2025hademif,li2025multi,xue2024unifying}. Xin et al.~\cite{xin2025lumina2,xin2025lumina} who introduced scalable autoregressive and omni-diffusion architectures that can be further used in advanced image editing. Liu et al.~\cite{liu2024difflow3d} enhanced scene flow estimation via iterative diffusion, while Zhang et al.~\cite{zhang2024synergistic} demonstrated the potential of diffusion processes in molecular and structural synthesis. Other representative efforts~\cite{ni2025wonderfree,xue2025similarity} continue to enrich multimodal benchmarks and architectures. It is noted that intention-driven visual reasoning~\cite{chen2025visrl} further reveal the emerging synergy between structured reasoning and perceptual modeling, inspiring the future direction of neural-driven methods toward reasoning models. 

Collectively, these works point a bright future research path, which leverages diffusion-driven priors, causal reasoning, and behaviorally grounded learning for more interpretable and controllable multimodal intelligence based on advanced neural-driven approach and updated BCI devices.

\section{Supplementary Dataset Details}
\label{dataset}

\subsection{Background Details}

\textbf{Overall Mechanism.} As illustrated in Fig.~\ref{fig:collectMach}, EEG captures electrical activity along the scalp with high temporal resolution, enabling fine-grained monitoring of neural dynamics. Neural signals from Fpz, Fp2, Pz, and Oz electrodes serve complementary roles in facilitating hands-free image editing. Fpz and Fp2, located in the frontal cortex, are primarily responsible for attentional control and intentional decision-making, respectively. Notably, we deliberately preserve blink-related signals at Fpz and Fp2, allowing the system to retain ecologically valid user states and incorporate implicit ocular cues without requiring additional eye-tracking hardware. Pz supports spatial attention and target selection, while Oz reflects visual perception and validation of image modifications.
Importantly, we avoid relying on conventional motor imagery regions (e.g., C3/C4), which are associated with imagined limb movements and commonly used in traditional BCI paradigms. These approaches often require extensive training, suffer from high inter-subject variability, and may be inaccessible to users with motor impairments. Instead, LoongX prioritizes frontal and parietal EEG channels that reflect universal cognitive processes such as attention, intention, and visual evaluation. This design enables plug-and-play usability without the need for motor calibration, ensuring a more intuitive, low-burden, and inclusive experience across diverse user populations.
To further enhance the decoding of user states, we incorporate fNIRS signals from the left and right anterior medial prefrontal cortex (aMPFC), which provide critical hemodynamic insights into cognitive load, emotional valence, and motivation intensity. In addition, peripheral PPG signals are used to capture autonomic responses such as heart rate variability and arousal levels, enabling the system to monitor stress and engagement during editing tasks. It offers complementary physiological information, such as heart rate, blood oxygen saturation (SpO2), and peripheral blood flow variations. 
Given the susceptibility of these biosignals to motion-induced artifacts, particularly those arising from head movements, a triaxial accelerometer and a triaxial gyroscope are employed to capture translational and rotational motion of the head. This motion data is subsequently used for signal quality assessment and artifact mitigation, thereby enhancing the reliability of the acquired physiological measurements.
Overall, this multimodal integration allows LoongX to decode user intent, attention, and emotional context in a holistic and adaptive manner, resulting in more precise, reliable, and user-aware editing commands.

\subsection{Subject Information}
\label{subsec:subjectinformation}
12 healthy college students (6 females and 6 males) were recruited as the subjects for data collection. They have a mean age of 24.5$\pm$2.5 years and normal (or corrected-to-normal) vision. All volunteers were informed of the experimental process and received financial compensation. All volunteers signed the consent forms prior to the experiment, which was approved by the Ethics Committee of the ZJU Review Board for Human Research Protections (Protocol ID: No. 067 (2019)). 
The attention score in this study is objectively computed using the ratio between the power of the EEG alpha band (8--12 Hz) and the theta band (4--8 Hz), as shown below:
\begin{equation}
\text{Attention Score} = \frac{\text{Alpha Band Power}}{\text{Theta Band Power}}
\end{equation}

This ratio has been widely used in cognitive neuroscience research as a neurophysiological index of attentional control and mental workload. For instance, Raufi and Longo~\citep{raufi2022evaluating} demonstrated that the alpha-to-theta and theta-to-alpha band ratios are reliable indicators of self-reported mental workload levels in EEG-based studies. The attention scores and the neural-signal-based image editing errors of 15 subjects in our experiment are summarized in Table~\ref{tab:tab_individual_score}. Subjects 13-17 are regarded as unseen subjects, and the L-Mind training set does not include these unseen data for cross-subject evaluation.

\begin{table*}[ht]
\centering
\caption{Attention and neural-signal-based image editing scores of individual subject}
\begin{adjustbox}{max width=\textwidth}
\begin{tabular}{lcccccccccc}
\toprule
\textbf{Subject} & \textbf{Gender} & \textbf{Age} & \textbf{\#Samples} & \textbf{Attention} & \textbf{L1} & \textbf{L2} & \textbf{CLIP-I} & \textbf{DINO} & \textbf{CLIP-T} \\
\midrule
1  & Female & 25 & 2003 & 0.0887 & 0.2657 & 0.1109 & 0.6370 & 0.4890 & 0.2196 \\
2  & Female & 29 & 2000 & 0.0817 & 0.2416 & 0.0950 & 0.6575 & 0.5021 & 0.2249 \\
3  & Female & 26 & 2001 & 0.1340 & 0.2448 & 0.0963 & 0.6660 & 0.4878 & 0.2337 \\
4 & Female & 22 & 1999 & 0.0739 & 0.2533 & 0.1005 & 0.6394 & 0.4606 & 0.2270 \\
5 & Female & 28 & 1992 & 0.1218 & 0.2552 & 0.1031 & 0.6144 & 0.4157 & 0.2260 \\
6 & Female & 29 & 1964 & 0.0822 & 0.2511 & 0.1000 & 0.6449 & 0.4564 & 0.2213 \\
7  & Male   & 22 & 1988 & 0.0851 & 0.2711 & 0.1160 & 0.6515 & 0.4634 & 0.2234 \\
8  & Male   & 23 & 1993 & 0.1105 & 0.2528 & 0.1017 & 0.6638 & 0.4833 & 0.2242 \\
9  & Male   & 22 & 1988 & 0.1500 & 0.2497 & 0.0998 & 0.6355 & 0.4571 & 0.2212 \\
10  & Male   & 24 & 2000 & 0.1298 & 0.2657 & 0.1144 & 0.6194 & 0.4240 & 0.2220 \\
11  & Male   & 24 & 2000 & 0.0954 & 0.2744 & 0.1151 & 0.6386 & 0.4339 & 0.2299 \\
12  & Male   & 22 & 2000 & 0.0971 & 0.2551 & 0.1034 & 0.6213 & 0.4323 & 0.2250 \\
\midrule
13 (unseen) & Male & 35 & 500 & 0.1210 & 0.2681 & 0.1174 & 0.6022 & 0.4418 & 0.2594 \\
14 (unseen) & Female & 30 & 500 & 0.0775 & 0.2688 & 0.1179 & 0.6051 & 0.4405 & 0.2553 \\
15 (unseen) & Male & 13 & 200 & 0.0727 & 0.2618 & 0.1001 & 0.6055 & 0.4472 & 0.2576 \\
16 (unseen) & Female & 62 & 100 & 0.0441 & 0.2660 & 0.1141 & 0.6196 & 0.4611 & 0.2472 \\
17 (unseen) & Male & 63 & 100 & 0.0520 & 0.2610 & 0.1133 & 0.6017 & 0.4588 & 0.2595 \\
\bottomrule
\end{tabular}
\label{tab:tab_individual_score}
\end{adjustbox}
\end{table*}

\subsection{Data Collection Details}
Using the setup shown in Fig.~\ref{fig:collect}, a total of 23,928 pairs of effective data were collected from 12 participants.
The specific details are given as follows.

\textbf{Multimodal Device and Signal Collection Pipeline.}
LoongX employs non-invasive BCI technologies to acquire multimodal neurophysiological signals, integrating data from four EEG channels (Fpz, Fp2, Pz, Oz) sampled at 250 Hz, eight fNIRS channels located in the medial prefrontal cortex (MPFZ) zone sampled at 25 Hz, and eight PPG channels also within the MPFZ zone sampled at 25 Hz. Additionally, six channels of head motion data—comprising triaxial linear acceleration and triaxial angular velocity—are recorded at 12.5 Hz to capture head movement dynamics. EEG, fNIRS, PPG, and motion signals are synchronized and transmitted from the device to the PC-side software application via Bluetooth 5.3. The application streams these signals in real time to the Lab Recorder software through the Lab Streaming Layer (LSL). Meanwhile, event markers generated by the image-stimulus paradigm on the PC are also sent via LSL to Lab Recorder. These markers indicate the start and stop times of user-triggered recordings. As a result, we obtain well-aligned EEG, fNIRS, PPG, motion, and audio signals. 
There was a 1-second cross-fixation interval between each pair of pre- and post-edited images (Fig.~\ref{fig:collect}). Data collection was organized in batches of 100 images, after which participants were given a rest period. A total of 2,000 images were collected from each participant. A total of 23,928 pairs of effective data were collected, comprising participants' speech, EEG, fNIRS, PPG, and head motion information. 

Our developed computer software used in this non-invasive BCI device streams the data in real time to the \textit{Lab Recorder} software through the Lab Streaming Layer (LSL) framework. Meanwhile, event markers generated by the image-stimulus paradigm on the PC are also sent via LSL to \textit{Lab Recorder}. These markers indicate the start and stop times of user-triggered recordings. Neural signals are collected simultaneously while the participant uses speech to describe the content of the image editing. As a result, we obtain well-aligned EEG, fNIRS, PPG, motion, and audio signals. There was a 1-second cross-fixation interval between each pair of pre- and post-edited images to distinguish between different image-editing pairs clearly. 2,000 images are collected from each participant. In all, 23,928 valid data pairs are gathered, encompassing participants' speech, EEG, fNIRS, PPG, and head motion information.

\textbf{Participants.} 12 healthy college students (6 female and 6 males) were recruited as the subjects for data collection. They have a mean age of 24.5$\pm$2.5 years and normal (or corrected-to-normal) vision. All volunteers were informed of the experimental process and received financial compensation (at an hourly rate exceeding the minimum wage in the region). All volunteers signed the consent forms prior to the experiment, which was approved by the Ethics Committee of the Institutional Review Board at Zhejiang University for Human Research Protections (Protocol ID: No. 067 (2019)). 
To ensure data consistency during extended EEG acquisition, the experiment was conducted in a quiet room (3m × 5m) maintained at a constant temperature of 24°C and a constant humidity. EEG signals were recorded using the latest non-invasive hydrogel electrodes, which are known to provide one of the highest signal quality among available electrode types. To maximize signal integrity and minimize impedance-related artifacts, the electrodes were replaced every five hours or sooner if necessary. Data acquisition for each participant starts at 9 AM daily, and the room was shielded from direct sunlight to reduce its impact on light-sensitive signals such as fNIRS and PPG. 
Example stimulus sessions that were displayed to participants are shown in Fig.~\ref{fig:collect}.

\subsection{Data Preprocessing}

To make use of the most relevant information and reduce noise and artifacts, each type of multimodal neurophysiological signal was preprocessed based on its unique characteristics. The first step for our proposed neural-based image editing method is extracting and encoding data into a structured format based on collected metadata.
Techniques such as interpolation for missing data and normalization methods to standardize signal amplitudes are applied to clean and normalize data to remove noise, fill in missing values, or correct errors.
Advanced filtering skills based on machine learning are implemented to select data aligned with the model’s predefined objectives, such as identifying visual patterns correlated with specific cognitive states or processing visual-related neural signals. Specifically, the data processing steps include:

\textbf{EEG Preprocessing.} Signals were band-pass filtered (1–80 Hz) and notch-filtered (48–52 Hz) to remove noise and powerline artifacts. The EEG channels near the eyes (Fpz and Fp2) retained ocular signals for intentional blink detection. Specific procedures are as follows:
\begin{itemize}
  \item A band-pass filter was applied in the range of 1–80 Hz to remove low-frequency drifts and high-frequency noise.
  \item A notch filter was applied in the frequency range of 48–52 Hz to eliminate powerline interference, which is specific to the 50 Hz electrical grid in China.
  \item Ocular artifacts from the AF8 and FPz channels were preserved intentionally, as these channels were specifically designed to capture eye movements, including blinking.
\end{itemize}

\textbf{fNIRS Preprocessing.}
The eight-channel fNIRS signals were processed to extract concentration changes of oxygenated hemoglobin (HbO), deoxygenated hemoglobin (HbR), and total hemoglobin (HbT). The following preprocessing steps were performed:
\begin{itemize}
  \item The HbO, HbR, and HbT signals were band-pass filtered in the range of 0.01–0.5 Hz to isolate brain spontaneous and stimulus-induced hemodynamic responses associated with neural activity. These signals correspond to brain activity evoked by the image-editing paradigm.
  \item The processed HbO, HbR, and HbT signals were averaged to obtain two signals per hemisphere: left hemisphere HbO, right hemisphere HbO, left hemisphere HbR, right hemisphere HbR, left hemisphere HbT, and right hemisphere HbT.
  \item Moreover, light intensities at 735 and 850 nm were converted to HbO/HbR/HbT using the Modified Beer–Lambert Law, then filtered (0.01–0.5 Hz) to isolate hemodynamic responses.
\end{itemize}

\textbf{Hemodynamic Conversion}: fNIRS signals are calculated by raw 735 nm and 850 nm near-infrared light intensity. The raw fNIRS signals are measured as light intensity changes at different wavelengths. To convert these optical signals into concentrations of oxygenated hemoglobin (HbO) and deoxygenated hemoglobin (HbR), we apply the Modified Beer–Lambert Law (MBLL). The logarithmic optical density change $\Delta A$ is calculated as:

\begin{equation}
\Delta A(\lambda) = \log\left( \frac{I_0(\lambda)}{I(\lambda)} \right)
\end{equation}

where $I_0(\lambda)$ is the initial light intensity, $I(\lambda)$ is the measured intensity at wavelength $\lambda$, and $\Delta A(\lambda)$ is the optical density change.

Then, the concentration changes of HbO and HbR are derived using the MBLL as follows:

\begin{equation}
\begin{bmatrix}
\Delta \text{HbO} \\
\Delta \text{HbR}
\end{bmatrix}
= \frac{1}{\text{DPF} \cdot L}
\cdot
\begin{bmatrix}
\varepsilon_{\text{HbO}}^{\lambda_1} & \varepsilon_{\text{HbR}}^{\lambda_1} \\
\varepsilon_{\text{HbO}}^{\lambda_2} & \varepsilon_{\text{HbR}}^{\lambda_2}
\end{bmatrix}^{-1}
\cdot
\begin{bmatrix}
\Delta A(\lambda_1) \\
\Delta A(\lambda_2)
\end{bmatrix}
\end{equation}

Here, $\varepsilon$ denotes the molar extinction coefficients of HbO and HbR at wavelengths $\lambda_1$ and $\lambda_2$, $L$ is the source-detector separation, and $\text{DPF}$ is the differential pathlength factor that accounts for scattering in tissue.
Additionally, the total hemoglobin concentration change $\Delta \text{HbT}$ is calculated as the sum of the changes in HbO and HbR:

\begin{equation}
\Delta \text{HbT} = \Delta \text{HbO} + \Delta \text{HbR}
\end{equation}

This process yields time-series concentration changes $\Delta$HbO, $\Delta$HbR, and $\Delta$HbT. These eight channels of $\Delta$HbO, $\Delta$HbR, and $\Delta$HbT were band-pass filtered (0.01–0.5 Hz) to isolate brain hemodynamic responses related to neural activity evoked by the image-editing paradigm. Finally, the HbO, HbR, and HbT signals were then averaged to obtain left and right hemisphere values for each.

\textbf{PPG Preprocessing.}
Optical signals were band-pass filtered (0.5–4 Hz) to extract cardiac rhythms for heart rate variability and arousal estimation. The raw light intensity data for the four-channel PPG signals were acquired at wavelengths of 735 nm and 850 nm. The detailed preprocessing steps for these signals were as follows:
\begin{itemize}
  \item The raw optical intensity data at both 735 nm and 850 nm were averaged to obtain the following four signals: left hemisphere 735 nm, right hemisphere 735 nm, left hemisphere 850 nm, and right hemisphere 850 nm.
  \item A band-pass filter with a frequency range of 0.5–4 Hz was applied to these four signals to extract hemodynamic changes associated with cardiac rhythms, reflecting the blood flow pulsations due to heartbeats.
\end{itemize}

\textbf{Motion Tracking.} The raw signals from the six-axis motion sensors (accelerometer + gyroscope) were retained to ensure accurate monitoring of head movements during the experiment, as these movements could potentially affect signal quality. These signals were also used to assess signal quality and reduce movement-related artifacts.

\textbf{Session Segmentation.}: Each editing session was defined by a speech-triggered onset and offset, linked to pre- and post-edited image pairs. Only active segments were retained for training and testing. Specifically, the image stimulus is used to mark the beginning of a session when the user starts recording audio. The session ends when the user stops the audio recording. During this time, the system synchronizes the recording of the participant's speech, EEG, fNIRS, PPG, and head motion data. Data recorded during periods of silence, when the audio recording is not active, is excluded from training and analysis.

%\textbf{Labeling and Quality Checking}: Each session was labeled based on the corresponding paired image name.
%The raw EEG and fNIRS signals underwent a series of final checking steps to ensure data quality and reliability: 1) Artifact removal: Techniques such as Independent Component Analysis (ICA) were applied to identify and eliminate artifacts arising from eye movements, muscle activity, and environmental noise; 2) Feature analysis: From the cleaned signals, features were extracted relevance analysis. For EEG, it ensured power spectral densities within specific frequency bands (e.g., alpha, beta, theta rhythms) and event-related potentials (ERPs). For fNIRS, features such as changes in oxygenated and deoxygenated hemoglobin concentrations were analyzed to ensure completeness.

\section{Supplementary Method Details}

\subsection{Background Information}

To enable hands-free image editing, the integration of multimodal signals facilitates the interpretation of the user’s intent and mental state: EEG activity at frontal sites such as Fpz reflects attention control and task initiation~\citep{adjouadi2004interpreting}; signals from the right prefrontal cortex (Fp2) are associated with emotional regulation and motivational drive, as evidenced by frontal alpha asymmetry~\citep{constant2012eeg}; EEG at the midline parietal site (Pz) captures spatial attention and visuomotor integration~\citep{babiloni2004abnormal}; and occipital EEG (Oz) provides information on visual perception and image processing load~\citep{adcock2012occipital}. Additionally, fNIRS measurements over the left and right anterior medial prefrontal cortex (aMPFC) reveal cognitive load and emotional valence through blood oxygenation patterns~\citep{Si2023}, and peripheral photoplethysmography (PPG) signals monitor heart rate variability as a proxy for autonomic arousal, enabling the system to track user stress levels and engagement during interaction~\citep{Arsalan2021,Tazarv2021}. Integrating these modalities allows the system to adapt to the user’s focus, emotional state, and workload, making hands-free image editing more intuitive and responsive.

Midline parietal EEG (fpz) is involved in spatial attention allocation and visuomotor integration, contributing to the coordination of visual and motor aspects of the editing task \cite{adjouadi2004interpreting}. Occipital EEG at Oz captures visual perception dynamics and image processing load, providing insight into how visual information is being processed by the user’s brain \cite{adcock2012occipital}.

%LoongX utilizes non-invasive BCI technologies to capture neurophysiological signals, specifically 4 channels (250 Hz EEG AF8, Pz, Oz, Fpz), 8 channels (MPFZ zone) 25 Hz fNIRS, 8 channels (MPFZ zone) 25 Hz PPG, combined with 6 channels 12.5 Hz head motion information (X, Y, and Z axial acceleration, rotational angular velocity about X, Y, and Z axis). As shown in Figure~\ref{fig:collect}, EEG measures electrical activity along the scalp, providing high temporal resolution of neural dynamics, while fNIRS assesses hemodynamic responses associated with neural activity, offering insights into cortical oxygenation levels. Furthermore, the PPG signal contains information such as heart rate, SpO2, and blood flow changes. Together, these modalities enable a comprehensive understanding of the user's cognitive states during image editing tasks. 

%In addition, these biosignals are easy to receive interference from head movement, resulting in a great decrease in signal credibility, so we use 3-axis accelerometer and 3-axis gyroscope to record the direction of head translation and rotation for signal quality detection.

\subsection{Theoretical Derivation of Flow-Aware Inversion}

We defined inversion as a trajectory that transports a clean sample $x_{0}\!\sim\!p_{0}(x)$ to a noisy latent $x_{t}\!\sim\!p_{t}(x)$.  Within the DDPM framework, the forward process is described as:

$$
x_{t}= \sqrt{\bar\alpha_{t}}\,x_{0}\;+\;\sqrt{1-\bar\alpha_{t}}\;\varepsilon,\qquad
\bar\alpha_{t}=\!\prod_{i=1}^{t}\alpha_{i},\;\;\alpha_{i}=1-\beta_{i},\;\;
\varepsilon\sim\mathcal N(0,I).
$$

First, we formulate a pure stochastic SDE that follows the forward diffusion to gradually add noise, and then run the time-reversed SDE to retrieve an editable reconstruction, similar to the philosophy of SDEdit~\cite{mengsdedit}.

Second, a probability-flow ODE treats diffusion via the score-based velocity field $v(x_\tau)$, replacing the random noise with a deterministic velocity field $v(x_{\tau})$ proportional to the score $\nabla_{x_{\tau}}\log p_{\tau}(x_{\tau})$:

$$
x_{0}=x_{t}-\int_{0}^{t}v(x_{\tau})\,\mathrm d\tau,\qquad
x_{t}=x_{0}+\int_{0}^{t}v(x_{\tau})\,\mathrm d\tau
      =x_{0}-\int_{t}^{0}v(x_{\tau})\,\mathrm d\tau.
$$

A continuum between these two extremes is obtained by interpolating the stochastic and deterministic contributions with a parameter $\eta\in[0,1]$:

$$
x_{t}= \sqrt{\bar\alpha_{t}}\,x_{0}
      +\sqrt{1-\bar\alpha_{t}}\,
        \bigl[\eta\,\varepsilon + (1-\eta)\,u_{t}\bigr],
\quad
u_{t}= \int_{0}^{t}\frac{c_\tau\,v(x_{\tau},\tau)}{\sqrt{1-\bar\alpha_{\tau}}}\,\mathrm d\tau ,
$$
where $\varepsilon \sim \mathcal{N}(0, I)$ and $c_\tau$ is a schedule-dependent factor that aligns the units of the velocity term with standard DDPM dynamics~\cite{ho2020denoising}. Choosing $\eta = 0$ recovers the deterministic ODE path, whereas $\eta = 1$ yields the fully stochastic SDE path, and intermediate values trade deterministic guidance for stochasticity.

Our flow-aware inversion belongs to the deterministic end.  As Flux.1-Dev predicts rectified-flow velocity rather than a DDPM score, we insert a lightweight rank-128 LoRA adapter $W$ that maps the frozen backbone’s predicted velocity $\boldsymbol\epsilon_{\phi}(x_{\tau},\tau)$ into the DDPM score domain through:

$$
v(x_{\tau}) = \sigma_{\tau}W\bigl(\boldsymbol\epsilon_{\phi}(x_{\tau},\tau)\bigr).
$$

The time-dependent coefficient $\sigma_\tau$ helps bridge the rectified-flow velocity and the DDPM score scale, while the linear bridge preserves the benefits of flow pre-training and enables faithful one-to-one reconstructions, in a similar spirit to edit-friendly DDPM~\cite{huberman2024edit} or LEDITS++~\cite{brack2024ledits++}.

\subsection{More Method Explanations}

To validate neural signals as reliable semantic conditions, we conduct a preliminary multi-label classification experiment detailed in Fig.~\ref{fig:classification}. Results show that EEG signals yield over 7\% mean Average Precision (mAP) gain compared to random noise, and fNIRS improved recall through robustness. Combining EEG and fNIRS shows stronger performance, and integration with textual prompts further enhances outcomes than using text only, confirming modality complementarity.

To handle diverse input shapes of raw signals, we pad/truncate inputs and find that EEG sequences of length 8,192 offer a good trade-off between performance and efficiency, which is summarized in Fig.~\ref{fig:classification}(c). As longer input sequences bring unbearable computational costs, we seek an optimal solution that keeps key information in the long sequence data while not significantly increasing computational costs. This motivates our CS3 encoder, which captures both temporal and channel-wise patterns efficiently while achieving a better trade-off between information density and computing efficiency.

To effectively extract structured representations from diverse signals, we propose the CS3 encoder. CS3 utilizes the linear computational complexity and efficiency of the structured state space model (S3M)~\cite{guefficiently} for encoding long sequences into channel representations. Recognizing that increasing latent dimensions can still significantly raise computational costs, CS3 implements a cross-feature extraction mechanism that separately encodes temporal and channel information.

While the fixed latent dimension of the S3M can not fully capture the dynamic information in signals, we further apply an adaptive feature pyramid based on adaptive average pooling. Each signal is processed through modality-specific neural encoders to produce latent features. Taking a $C$-channel EEG sequence $\mathbf X \in \mathbb{R}^{C\times L}$ as an example, we first normalize it to $[-1,1]$ and pass it through two parallel S3M blocks capturing complementary dynamics.

To align and integrate multi-source signals, we further introduce DGF, which dynamically processes and fuses features across modalities. DGF can model inter-modality interaction modelling to form a unified latent space, which is also optionally usable for alignment with text embeddings to support a more hybrid conditioning.

In summary, CS3 captures multi-scale temporal and structural patterns in neural signals, consistent with findings that multi-band EEG features improve intent decoding. DGF performs selective multimodal fusion through dual gating, which follows prior successes of gating and normalization strategies in multimodal learning. On pairing EEG plus PPG with T5, and fNIRS plus Motion with CLIP. Rationale is also not arbitrary. T5 provides fine-grained token-level semantics that help precise instruction following, which complements the fast neural dynamics in EEG and the lightweight hemodynamics from PPG. CLIP provides robust global semantics that align with slower cortex-wide fNIRS signals and intentional head Motion.

\section{Supplementary Experimental Details}

\subsection{Cross-subject Experiment}

Our original dataset was collected from 12 participants (6 female, 6 male, mean age 24.5 ± 2.5 years), each contributing around 2,000 paired samples under carefully controlled experimental conditions. While our initial split ensured training/test separation, we acknowledge that the possibility of subject overlap could limit generalizability. To address this, we performed additional cross-subject evaluations with 5 new unseen participants (3 male, 2 female, ages 13–63, see Table~\ref{tab:tab_individual_score}). The results are presented in Table~\ref{tab:original_unseen_comparison}. It confirms that the model maintains strong generalization when applied to unseen individuals, with performance trends on CLIP-I, DINO, and CLIP-T remaining consistent with those from the original test set. This provides evidence that our approach is not overly reliant on subject-specific neural signatures, but instead captures transferable semantic representations.

\begin{table}[!htbp]
  \centering
  \small
  \vspace{-0.5em}
  \caption{Performance comparison of baseline and our proposed LoongX on the original test set (12 subjects) and unseen test set (5 new subjects).}
  \vspace{-0.5em}
  \renewcommand{\arraystretch}{1.25}
  \setlength{\tabcolsep}{1mm}
  \begin{tabular}{lccccccc}
    \toprule
    Test Dataset & Methods & Conditioning & L1 ($\downarrow$) & L2 ($\downarrow$) & CLIP-I ($\uparrow$) & DINO ($\uparrow$) & CLIP-T ($\uparrow$) \\
    \midrule[1.2pt]
    \multirow{3}{*}{Original} 
      & OmniControl & Text & 0.2632 & 0.1161 & 0.6558 & 0.4636 & 0.2549 \\
      & OmniControl & Speech & 0.2714 & 0.1209 & 0.6146 & 0.3717 & 0.2501 \\
      \rowcolor{gray!8}
      & LoongX (Ours) & Neural Signals & 0.2509 & 0.1029 & 0.6605 & 0.4812 & 0.2436 \\
      \rowcolor{gray!8}
      & LoongX (Ours) & Signals + Speech & 0.2594 & 0.1080 & 0.6374 & 0.4205 & 0.2588 \\
    \midrule
    \multirow{3}{*}{Unseen} 
      & OmniControl & Text only & 0.2581 & 0.1133 & 0.6528 & 0.4655 & 0.2553 \\
      & OmniControl & Speech & 0.2779 & 0.1271 & 0.6221 & 0.3942 & 0.2508 \\
      \rowcolor{gray!8}
      & LoongX (Ours) & Neural Signals & 0.2574 & 0.1090 & 0.6019 & 0.4037 & 0.2403 \\
      \rowcolor{gray!8}
      & LoongX (Ours) & Signals + Speech & 0.2668 & 0.1146 & 0.6049 & 0.4447 & 0.2568 \\
    \bottomrule
  \end{tabular}
  \vspace{-1em}
  \label{tab:original_unseen_comparison}
\end{table}

\subsection{Abalation Studies Breakdown}

Table~\ref{tab:brain-signal-results} and Table~\ref{tab:brain-channels} correspond to the detailed ablation studies illustrated in Fig.~\ref{fig:ablation_signals} and Fig.~\ref{fig:ablation_regions} in the main manuscript.

From a neuroscience perspective, the superiority of fNIRS and EEG integration aligns well with our understanding of brain physiology: EEG captures rapid electrical oscillations reflecting millisecond-level neuronal activity, while fNIRS provides complementary information about slow hemodynamic responses, reflecting regional brain activation. Their fusion exploits both temporal and spatial dynamics of cognition, which is particularly relevant for decoding the complex neural basis of visual and semantic processing required by image editing tasks. The limited effect of PPG and motion signals may stem from its primary focus on peripheral patterns, which are less directly involved in cortical information processing but still can affect the robustness of the performance.

Drilling down into Table~\ref{tab:brain-channels}, the channel-wise analysis offers intriguing support for established functional specialization in the human brain:

\begin{itemize}
    \item The Oz channel (occipital cortex) stands out in global image alignment and robustness metrics, mirroring its neuroanatomical role as the hub for early-stage visual perception. The occipital lobe, and especially the Oz electrode position, is known for processing visual stimuli, edge detection, and scene analysis. The strong performance observed here suggests that even in a data-driven, deep learning context, the fundamental dominance of the visual cortex in image-based tasks persists.
    \item In contrast, the Fpz channel (frontopolar cortex) exhibits heightened performance in metrics linked to semantic understanding and higher-order cognitive alignment. The prefrontal regions are responsible for executive functions such as attention, planning, and integrating multimodal information, which are essential for aligning generated content with textual or conceptual prompts.
\end{itemize}

These results not only validate long-standing neuroscientific theories, such as the hierarchical processing streams in the brain (from occipital “what is seen” to frontal “what does it mean/what to do”), but also provide practical guidelines: in settings where only a limited number of electrodes or sensors are available, prioritizing signals from functionally specialized regions (e.g., Oz for vision, Fpz for semantic or cognitive control) can maximize decoding efficiency for targeted tasks.

Furthermore, the convergence of these findings with classical brain science underscores the translational value of deep learning in cognitive neuroscience. It highlights the potential for future brain-computer interfaces to be not only data-driven but also “anatomy-aware,” leveraging our evolving map of the brain to design more effective and interpretable multimodal AI systems.

\begin{table}[ht]
\centering
\caption{Evaluation results for different signal combinations.}
\setlength{\tabcolsep}{1mm}
\begin{tabular}{lccccc}
\toprule
Metric                & Pure EEG & EEG + fNIRS & EEG + fNIRS + PPG & All Signals & All Signals + Text \\
\midrule
L1                    & 0.2641   & 0.2508      & 0.2631            & 0.2571                      & 0.2594                        \\
L2                    & 0.1078   & 0.1029      & 0.1123            & 0.1076                      & 0.1080                        \\
CLIP-I                & 0.5457   & 0.6604      & 0.6536            & 0.6274                      & 0.6374                        \\
DINO                  & 0.2963   & 0.4811      & 0.4942            & 0.4245                      & 0.4205                        \\
CLIP-T           & 0.2251   & 0.2436      & 0.2226            & 0.2481                      & 0.2588                        \\
\bottomrule
\end{tabular}
\label{tab:brain-signal-results}
\end{table}

\begin{table}[ht]
\centering
\caption{Results using different brain region signals. GT refers to ground truth. Ch means the channel.}
\setlength{\tabcolsep}{1mm}
\begin{tabular}{lcccccc}
\toprule
Condition          &  L1     & L2     & CLIP-I  & DINO   & CLIP-T (Ours) & CLIP-T (GT) \\
\midrule
EEG (All channels)        & 0.2508 & 0.1029 & 0.6604  & 0.4811 & 0.2436      & 0.2594     \\
EEG (Ch 0, Pz)  & 0.2509 & 0.1028 & 0.6486  & 0.4787 & 0.2314      & 0.2594     \\
EEG (Ch 1, Fp2)  & 0.2581 & 0.1070 & 0.6178  & 0.4150 & 0.2421      & 0.2594     \\
EEG (Ch 2, Fpz)  & 0.2486 & 0.1022 & 0.6669  & 0.4846 & 0.2481      & 0.2594     \\
EEG (Ch 3, Oz)  & 0.2475 & 0.1003 & 0.6619  & 0.4873 & 0.2367      & 0.2594     \\
\bottomrule
\end{tabular}
\label{tab:brain-channels}
\end{table}

% 0-3通道顺序分别为pz，fp2, fpz oz

\subsection{More Qualitative Results}
\label{sec:morecase}

\begin{figure*}[htbp]
	\begin{center}
		\includegraphics[trim=0cm 1cm 0 0, width=1\textwidth]{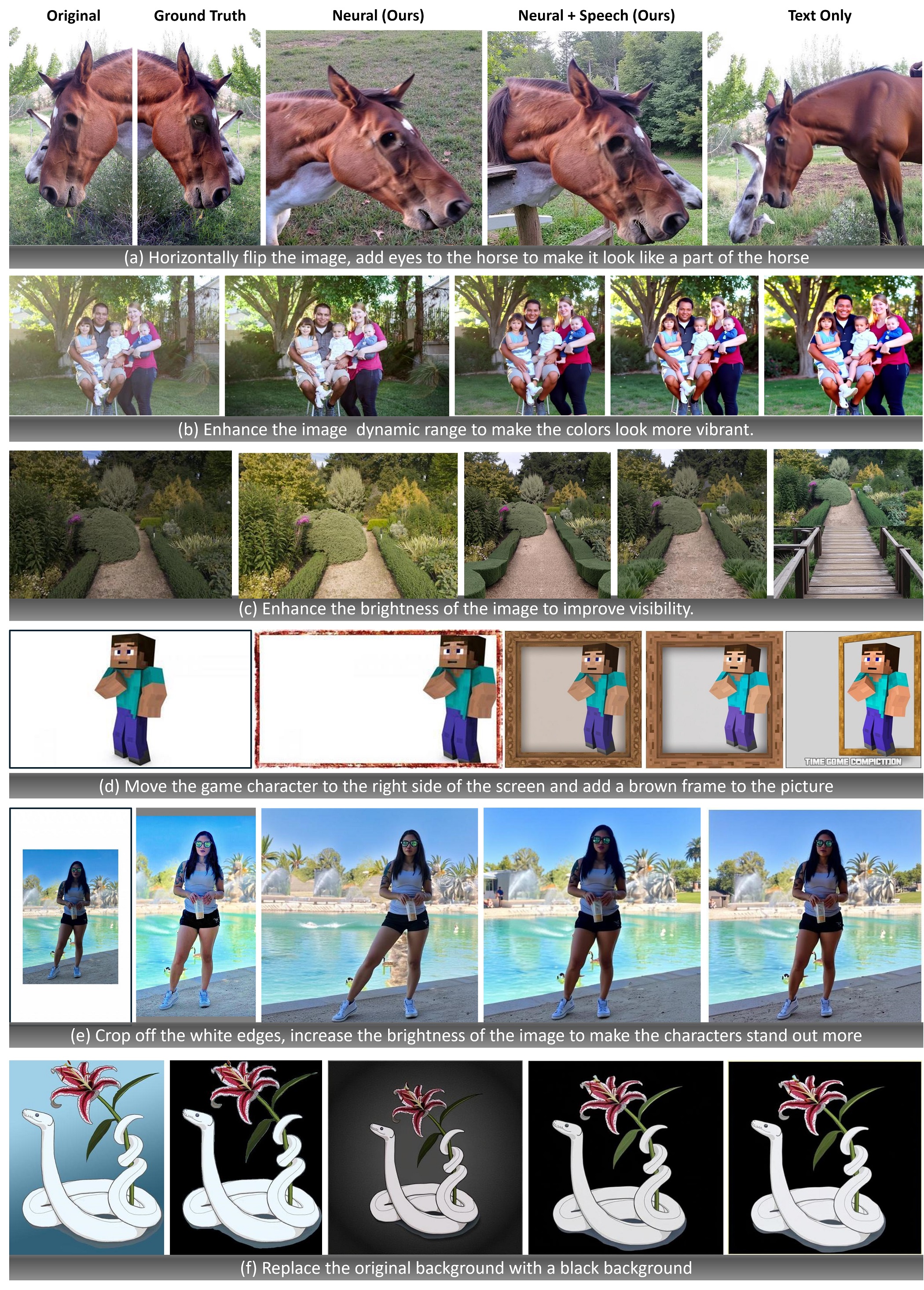}
        \vspace{0.5em}
		\caption{Qualitative comparison of our neural-driven and speech-neural fusion methods and text-prompt baseline for Global Editing category.}
		\label{fig:sup_case1}
	\end{center}
\end{figure*}

\begin{figure*}[htbp]
	\begin{center}
		\includegraphics[trim=0cm 1cm 0 0, width=1\textwidth]{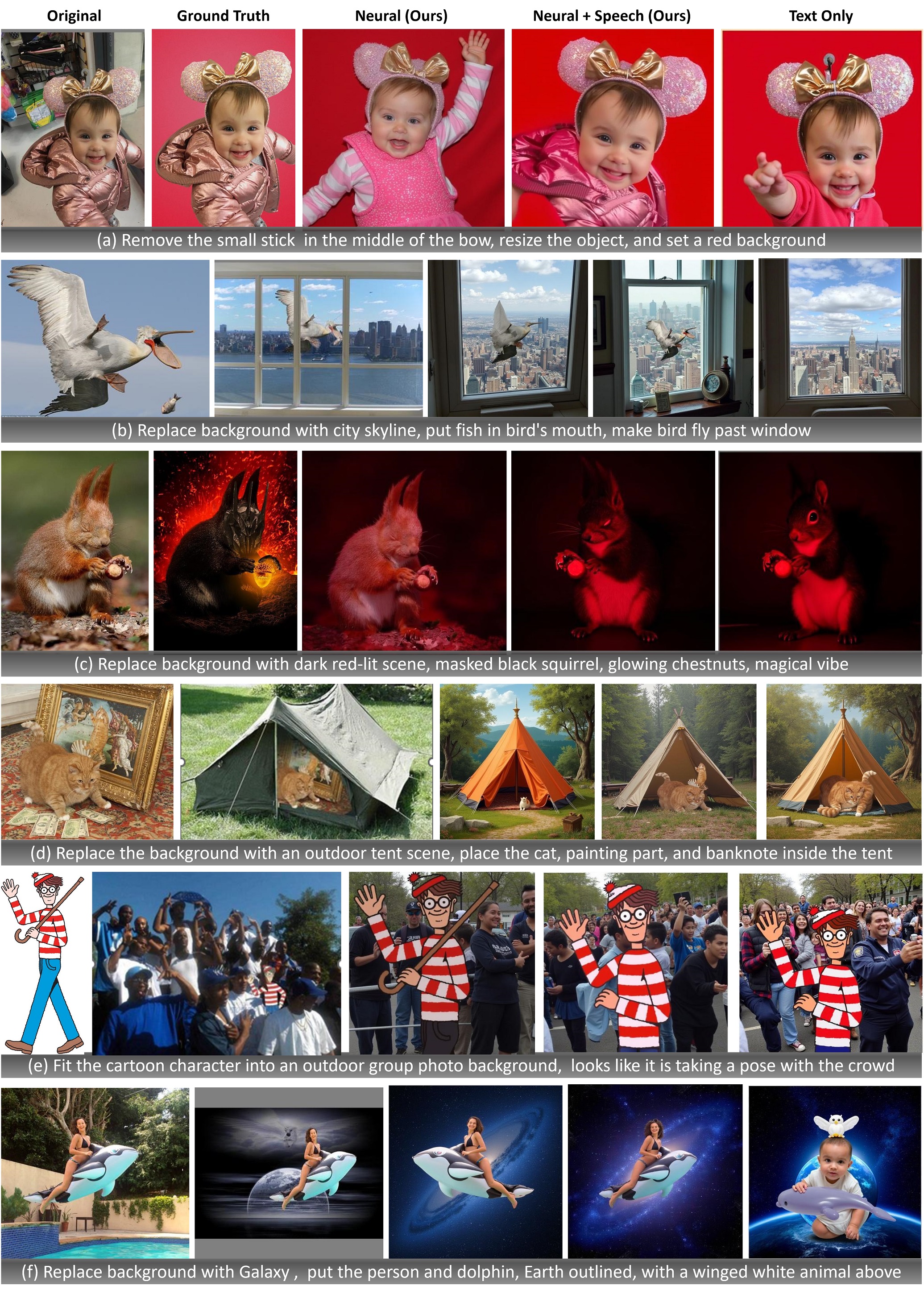}
        \vspace{0.5em}
		\caption{Qualitative comparison of our neural-driven and speech-neural fusion methods and text-prompt baseline for Background Editing category.}
		\label{fig:sup_case2}
	\end{center}
\end{figure*}

\begin{figure*}[htbp]
	\begin{center}
		\includegraphics[trim=0cm 1cm 0 0, width=1\textwidth]{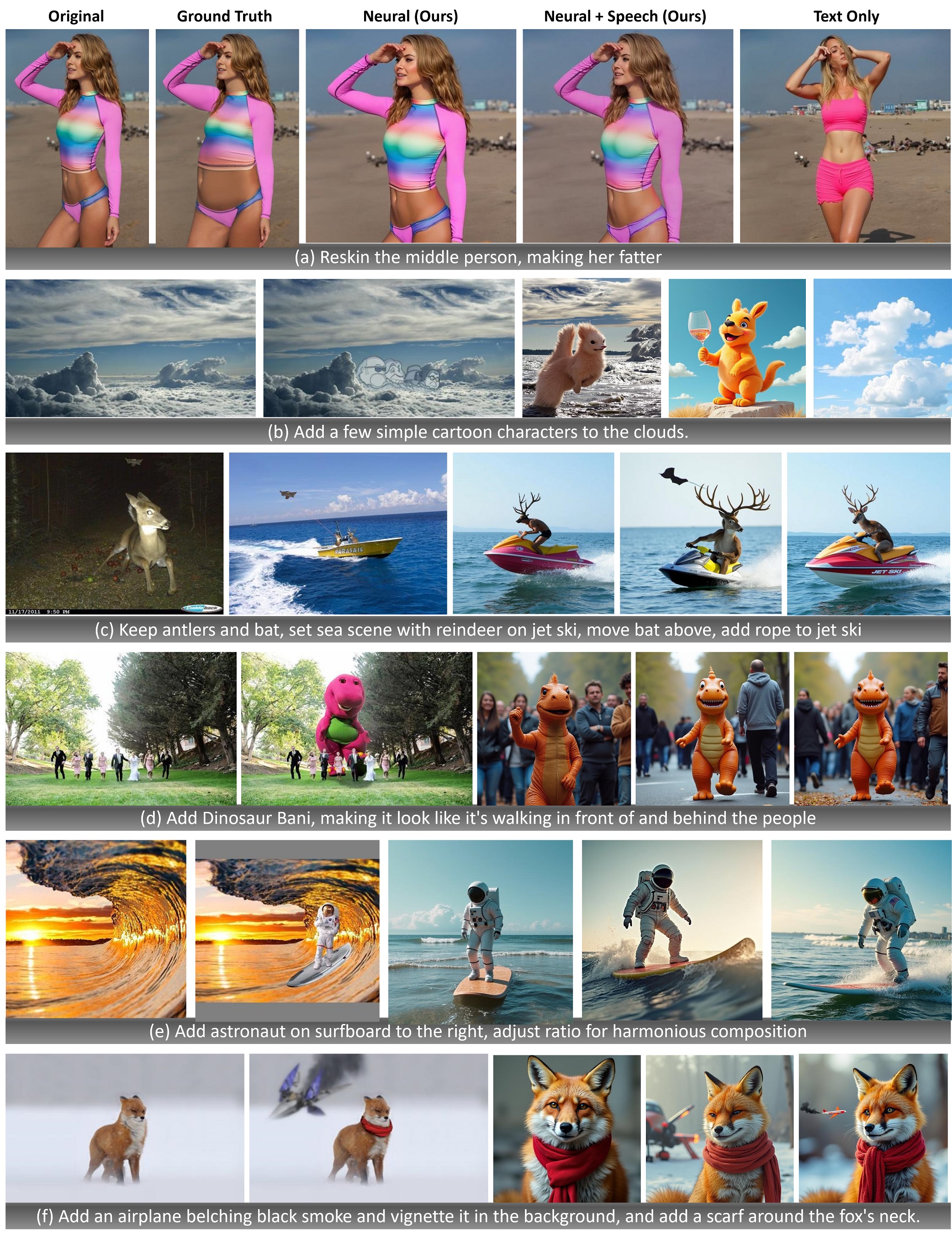}
        \vspace{0.5em}
		\caption{Qualitative comparison of our neural-driven and speech-neural fusion methods and text-prompt baseline for Object Editing category.}
		\label{fig:sup_case3}
	\end{center}
\end{figure*}

\begin{figure*}[htbp]
	\begin{center}
		\includegraphics[trim=0cm 2cm 0 0, width=1\textwidth]{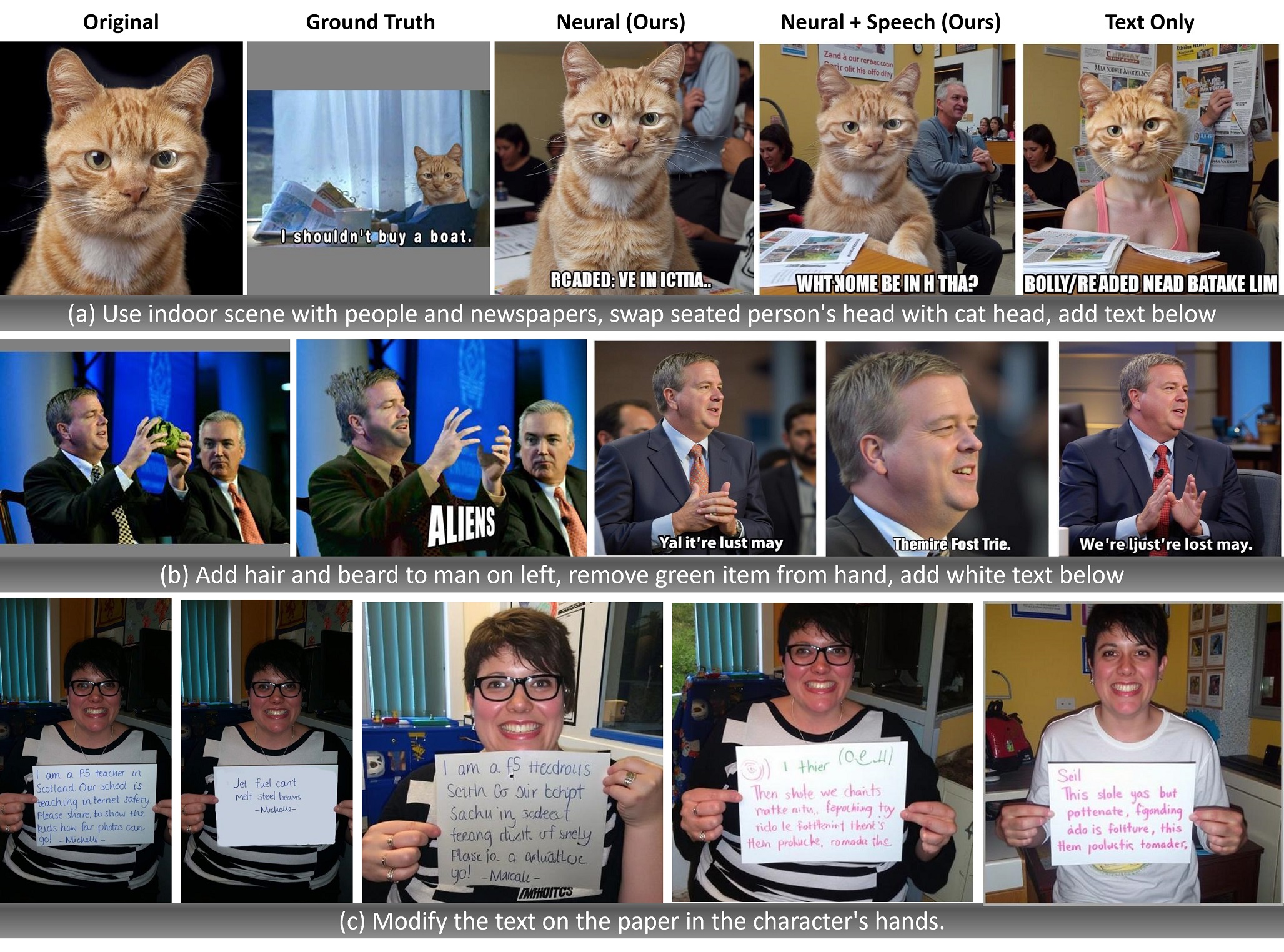}
		\caption{Qualitative comparison of our neural-driven and speech-neural fusion methods and text-prompt baseline for Text Editing category.}
        \vspace{-1.0em}
		\label{fig:sup_case4}
	\end{center}
\end{figure*}

More qualitative comparisons are presented in Figures \ref{fig:sup_case1}-\ref{fig:sup_case4}, corresponding to the four broad editing categories: Global, Background, Object, and Text Editing. For clarity and conciseness in the figures, the original lengthy instructions have been distilled into single-sentence descriptions without altering their intended meaning.
The editing results of our neural-driven and neural-speech fusion methods consistently outperform text-prompt-based editing results, demonstrating superior alignment with human intent and greater editing precision.
Notably, Text Editing presents a more complex challenge compared to other categories. Given the current limitations of backbone models (with the exception of commercial models like GPT-4o), text-based edits remain difficult. As evidenced by the examples, neural-driven approaches exhibit a stronger ability to align with human intent, making the editing process more intuitive and effective. It is foreseeable that in the near future, as reliable image-editing backbones become more accessible, neural-driven image editing will further stabilize and mature, evolving into an indispensable tool for everyday creative workflows.

Fig.~\ref{fig:sup_case5} specifically analyzes three characteristic failure modes: (1) cases involving overly imaginative descriptions that deviate significantly from the training data distribution (e.g., "long-legged space creature"), (2) ambiguous instructions with insufficient semantic details (particularly evident in case (b) where background retention specifications were omitted), and (3) challenges posed by non-standard input image dimensions (such as panoramic aspect ratios). These failure cases provide valuable insights into the current limitations of neural-based editing systems.

\subsection{More Failure Cases}

\begin{figure*}[htbp]
	\begin{center}
		\includegraphics[trim=0cm 2cm 0 0, width=1\textwidth]{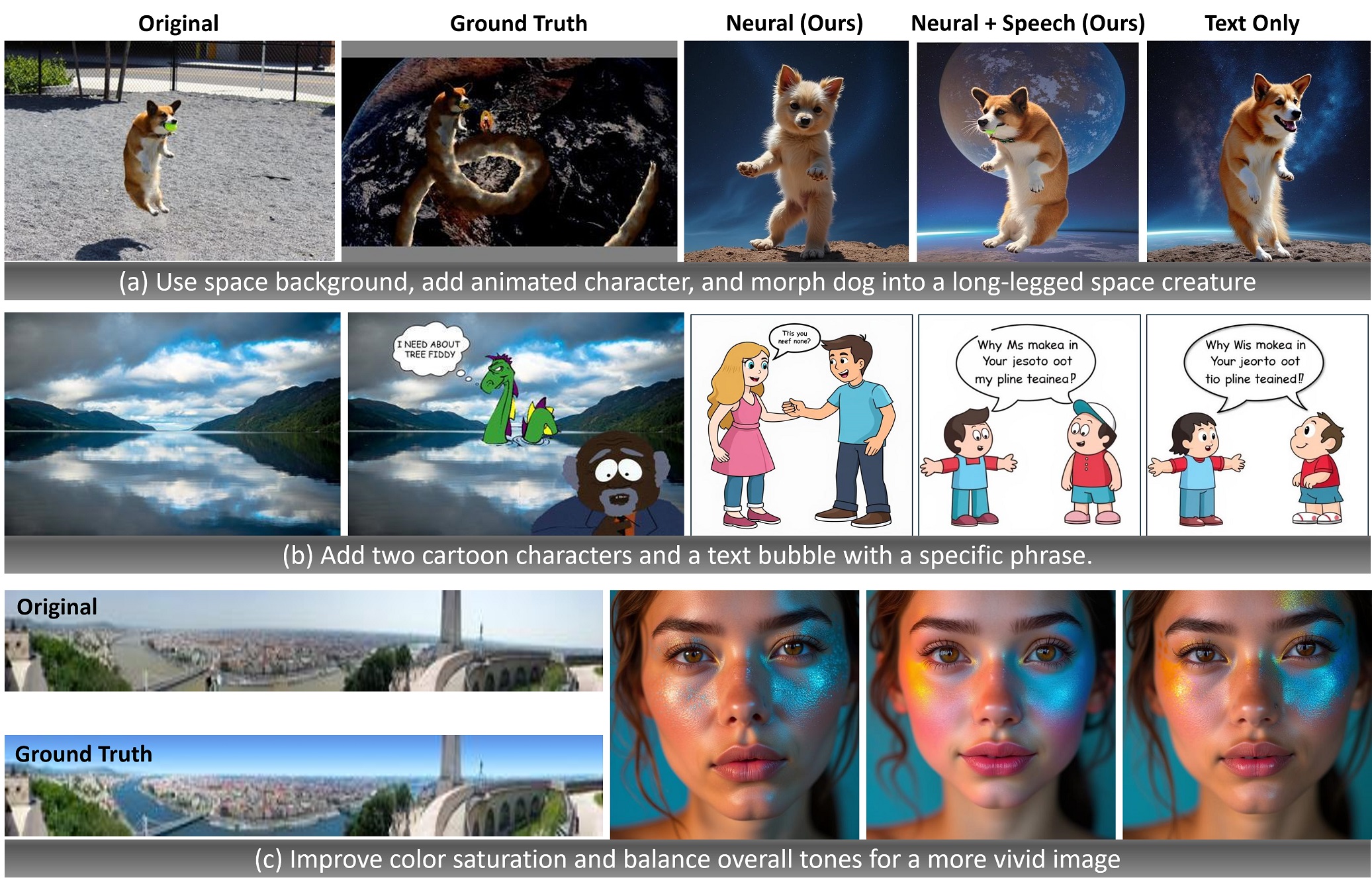}
		\caption{Qualitative analysiscomparison  onf our neural-driven and speech-neural fusion methods and text-prompt baseline for three failure cases: (a) Overly exaggerated descriptions, e.g., "long-legged space creature"; (b) Vague instructions lacking detail, such as omitting whether to retain the background; (c) Uncommon image dimensions, e.g., panoramic input images.}
		\label{fig:sup_case5}
	\end{center}
\end{figure*}

Figure. \ref{fig:sup_case5} specifically illustrates three failure cases, where overly exaggerated imagination, vague instructions, or uncommon input image sizes may contribute to failed results. It demonstrates three characteristic failure modes: (1) cases involving overly imaginative descriptions that deviate significantly from the training data distribution (e.g., "long-legged space creature"), (2) ambiguous instructions with insufficient semantic details (particularly evident in case (b) where background retention specifications were omitted), and (3) challenges posed by non-standard input image dimensions (such as panoramic aspect ratios). These failure cases provide valuable insights into the current limitations of neural-based editing systems.

\end{document}